\documentclass[11pt]{article}

\usepackage[preprint]{acl}

\usepackage{times}
\usepackage{latexsym}
\usepackage[T1]{fontenc}
\usepackage[utf8]{inputenc}
\usepackage{microtype}
\usepackage{inconsolata}
\usepackage{graphicx}

\usepackage{hyperref}
\usepackage{url}
\usepackage{booktabs}
\usepackage{amsfonts}
\usepackage{nicefrac}
\usepackage{xcolor}

\usepackage{amsmath}
\usepackage{amssymb}
\usepackage{mathtools}
\usepackage{amsthm}
\theoremstyle{plain}

\theoremstyle{definition}

\theoremstyle{remark}

\usepackage{bm}
\usepackage{graphicx}
\usepackage{subcaption}
\usepackage{float}
\usepackage{wrapfig}
\usepackage{array}
\usepackage{multirow}
\usepackage{dsfont}
\usepackage{cancel}
\usepackage{soul}
\usepackage[normalem]{ulem}
\usepackage{siunitx}
\usepackage{algorithm}
\usepackage{algorithmic}
\usepackage[page,toc,titletoc,title]{appendix}
\usepackage[capitalize,noabbrev]{cleveref}
\usepackage{colortbl}
\usepackage{tabularx}
\usepackage{tcolorbox}
\usepackage{mdframed}
\usepackage{pifont}
\usepackage{xargs}
\usepackage{cases}

\hypersetup{
	final,
	colorlinks,
	linkcolor=ourred,
	citecolor=ourblue,
	urlcolor=url,
}

\definecolor{ourblue}{rgb}{0.368,0.507,0.71}
\definecolor{ourorange}{rgb}{0.881,0.611,0.142}
\definecolor{ourgreen}{rgb}{0.56,0.692,0.195}
\definecolor{ourred}{rgb}{0.923,0.386,0.209}
\definecolor{ourviolet}{rgb}{0.528,0.471,0.701}
\definecolor{ourbrown}{rgb}{0.772,0.432,0.102}
\definecolor{ourlightblue}{rgb}{0.364,0.619,0.782}
\definecolor{ourdarkgreen}{rgb}{0.572,0.586,0.}

\definecolor{ourcyan2}{rgb}{0.125,0.722,0.804}
\definecolor{ourred2}{rgb}{0.863,0.184,0.047}
\definecolor{ouryellow2}{cmyk}{0,0.16,1.0,0.07}
\definecolor{ourviolet2}{cmyk}{0.55,0.56,0,0.47}
\definecolor{ourorange2}{cmyk}{0,0.46,0.89,0.11}

\definecolor{discretecolor}{RGB}{11,83,150}
\definecolor{gaussiancolor}{RGB}{230,145,56}
\definecolor{argmaxcolor}{RGB}{154,0,0}
\definecolor{grayseq}{RGB}{120,120,120}
\definecolor{url}{HTML}{c55a3a}
\definecolor{ourslightgray}{RGB}{235, 235, 235}

\newcommand{\Fig}[1]{Figure~\ref{#1}}
\newcommand{\fig}[1]{Fig.~\ref{#1}}

\newcommand{\tab}[1]{Table~\ref{#1}}
\newcommand{\Eqn}[1]{(\ref{#1})}
\newcommand{\eqn}[1]{eq.~\ref{#1}}

\renewcommand{\sec}[1]{Sec.~\ref{#1}}
\newcommand{\supp}[1]{Suppl.~\ref{#1}}

\newcommand{\algo}[1]{Algo.~\ref{#1}}

\newcommand{\mask}{[\textsc{mask}]~}

\newcommand{\barx}{\bar{\mathbf{x}}}

\usepackage{xspace}
\makeatletter
\DeclareRobustCommand\onedot{\futurelet\@let@token\@onedot}
\def\@onedot{\ifx\@let@token.\else.\null\fi\xspace}

\makeatother

\def\x{{\mathbf x}}

\def\supl{{\textcolor{grayseq}{\ensuremath{\ell}}}}
\def\prior{\boldsymbol{\pi}}
\newcommand{\Vsize}{|\mathcal{V}|}

\title{BlockGen: Flexible Blockwise Sequence Modeling with Hybrid Samplers}
\author{%
  Justin Deschenaux \\
  EPFL \\
  Lausanne, Switzerland \\
  \texttt{justin.deschenaux@epfl.ch} \\
  \And
  Caglar Gulcehre \\
  EPFL, Lausanne, Switzerland \\
  Microsoft AI \\
}

\begin{document}

\maketitle

\begin{abstract}
Is the uniform-state diffusion framework a more powerful paradigm for discrete diffusion? Recent studies indicate that this may be the case. In combination with predictor–corrector samplers, uniform-state diffusion models (USDMs) produce samples of higher-quality than masked diffusion models (MDMs), and USDMs equal or outperform MDMs in downstream tasks, even though they exhibit greater perplexity.
Two issues remain unresolved. First, existing work compares uniform and masked diffusion with un-informed correctors that re-inject noise at random positions, rather than targeting tokens most likely to be wrong. Second, prior work compares full-sequence diffusion models, so we do not know whether the same conclusion holds when tokens are generated block by block.
To address these issues, we introduce \emph{BlockGen}, a blockwise sequence model that we instantiate with both masked and uniform diffusion. BlockGen trains on a mixture of block sizes and its likelihood interpolates between AR and pure diffusion more finely than models with a fixed block size. BlockGen enables \emph{AR-informed predictor-corrector sampling} (ARPC), which combines AR and diffusion predictions to re-generate unlikely tokens without an auxiliary verifier.
Under ancestral sampling, uniform outperforms masked in the block-by-block setting, especially in the few-step regime. Under ARPC, the gap closes and reverses at high NFE. With block size $16$ on GSM8K, MDMs reach slightly higher accuracy than USDMs, and we observe a similar trend in Generative Perplexity on OpenWebText. Find our code at \url{https://github.com/jdeschena/blockgen}.
\end{abstract}
\section{Introduction}
Autoregressive (AR) Transformers are the dominant paradigm for sequence modeling and power modern language models through next-token prediction \citep{bengio2000neuralprobabilisticlanguagemodel, touvron2023llama2openfoundation, grattafiori2024llama3herdmodels, gemmateam2025gemma3technicalreport, openai2024gptoss}. The Transformer architecture \citep{vaswani2017attentionneed} has enabled training at scale, but AR generation is sequential and requires one forward pass per token, which limits throughput and latency. Causal attention can also hurt on reasoning tasks where bidirectional context is required \citep{papadopoulos2024arrowstimelargelanguage, kitouni2024factorizationcursetokenspredict, zhangli2024reversenumberdecodingorder, nagarajan2025roll}. Discrete diffusion language modeling \citep{sohldickstein2015deepunsupervisedlearningusing, austin2023structureddenoisingdiffusionmodels, campbell2022continuoustimeframeworkdiscrete, campbell2024generativeflowsdiscretestatespaces, gat2024discreteflowmatching, sahoo2024simpleeffectivemaskeddiffusion, shi2025simplifiedgeneralizedmaskeddiffusion, ou2025absorbingdiscretediffusionsecretly, lou2024discretediffusionmodelingestimating} is an alternative that iteratively refines a noisy sequence and can update many tokens per denoising step.

Recent comparisons between masked and uniform-state discrete diffusion favor USDMs. With ancestral samplers, uniform-state models match or surpass MDMs in downstream tasks despite worse perplexity \citep{sahoo2026scalingbeyondmasked}. USDMs also exhibit better test-time scaling under predictor-corrector sampling \citep{deschenaux2026diffusiondualitychapterii} and better scaling trends in the data-constrained regime \citep{vonrutte2025scalingbehaviordiscretediffusion}.
\begin{figure*}[t]
  \centering
  \begin{subfigure}[t]{0.49\textwidth}
    \centering
    \includegraphics[width=\textwidth]{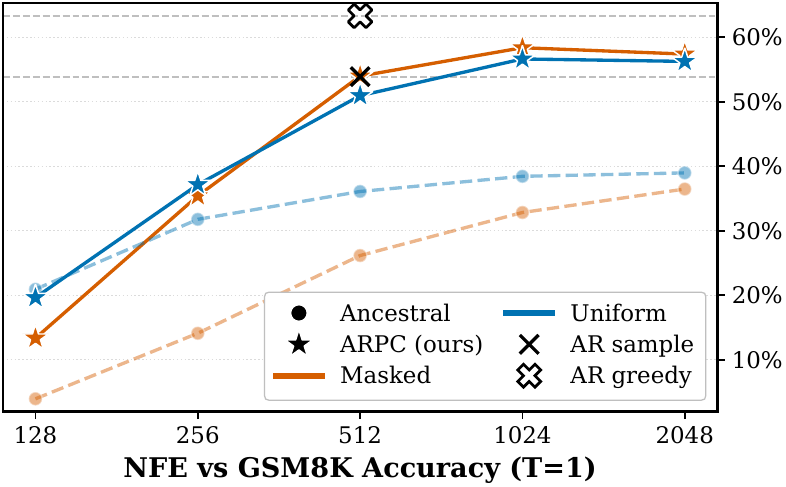}
    \label{fig:ar_vs_ancestral_vs_arpc_b16_T1}
  \end{subfigure}\hfill
  \begin{subfigure}[t]{0.49\textwidth}
    \centering
    \includegraphics[width=\textwidth]{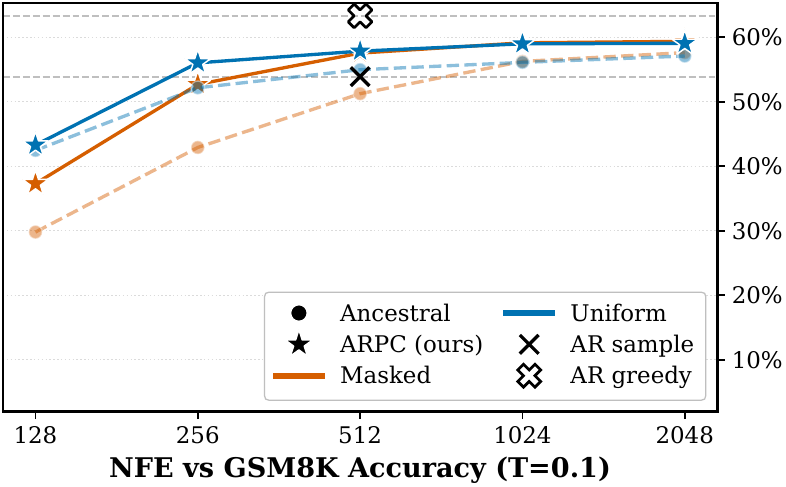}
    \label{fig:ar_vs_ancestral_vs_arpc_b16_T0p1}
  \end{subfigure}
      \caption{\textbf{GSM8K accuracy with block size 16} as a function of NFE (\emph{number of function evaluations}). Models are trained on TinyGSM and evaluated on the GSM8K test set. Each curve shows the best performance for a given NFE; the full sweep is in \supp{app:raw-accuracies}. \emph{AR-Informed Predictor-Corrector} (ARPC; ours) uses checkpoints trained with the mixture in \Eqn{eq:blockgen-mixture}, with $\gamma_1 = 0.05$, $\gamma_{16} = 0.95$. Ancestral uses a single-block-size model. Under ancestral sampling, uniform diffusion has higher accuracy than masked, with the largest gap at low NFE. Under ARPC the gap closes, and the masked-vs-uniform trend reverses at high NFE. ARPC also narrows the gap to AR with sampling. Greedy AR remains the strongest variant.}
  \label{fig:ar_vs_ancestral_vs_arpc_b16}
\end{figure*}

We investigate two open questions.
\textbf{(Q1) Block-by-block generation.} Block diffusion models \citep{arriola2025blockdiffusioninterpolatingautoregressive, wu2025fastdllmv2efficientblockdiffusion} generate tokens block-by-block, left-to-right, with block-causal attention that preserves the KV cache across decoded blocks. Prior work generally compares masked and uniform diffusion when the whole sequence is generated at the same time. Whether the advantage of uniform carries over when modeling sequences block-by-block is unclear.
\textbf{(Q2) Informed predictor-correctors.} Prior work compares uniform and masked with ancestral or \emph{un-informed} predictor-correctors samplers, which inject noise at random positions. Informed correctors that target likely mistakes \citep{zhao2025informedcorrectorsdiscretediffusion, liu2025thinkgeneratediscretediffusion, kim2025finetuningmaskeddiffusionprovable} were studied for masked diffusion but prior work generally does not compare MDMs with informed correctors to USDMs.
\paragraph{Contributions}
We introduce \emph{BlockGen}, a blockwise sequence model trained over a mixture of block sizes, instantiated with both masked and uniform diffusion within blocks. BlockGen admits a tractable ELBO that interpolates between AR and full-sequence diffusion.
(1) Mixture training improves perplexity over fixed-block BDMs on OpenWebText \citep{Gokaslan2019OpenWeb}.
(2) We design \emph{ARPC}, the \emph{AR-informed Predictor-Corrector} sampler, which uses the model's own AR predictions to score and re-generate unlikely tokens without an auxiliary verifier or extra training.
(3) In the block-by-block setting, USDMs outperform MDMs under ancestral sampling, especially in the few-step regime, so the full-sequence trend carries over (Q1). Under ARPC, the gap closes and reverses at higher NFE: on GSM8K \citep{cobbe2021trainingverifierssolvemath} at block size $16$ ($T{=}1$), masked + ARPC outperforms uniform + ARPC, and the Generative Perplexity on OpenWebText follows the same pattern (Q2).

\section{Background}
\label{sec:background}
\paragraph{Notation}
We represent the vocabulary as one-hot vectors $\mathcal{V} := \{\mathbf{v} \in \{0, 1\}^{\Vsize} : \|\mathbf{v}\|_1 = 1\}$. A sequence $\x \in \mathcal{V}^L$ consists of $L$ tokens, and $\x^\supl$ denotes its $\ell$-th element. We write $\Delta^{\Vsize}$ for the $\Vsize$-probability simplex and $\mathrm{Cat}(\cdot; \mathbf{v})$ for the categorical distribution with parameter $\mathbf{v} \in \Delta^{\Vsize}$. We denote by $\prior \in \Delta^{\Vsize}$ a fixed prior, $\mathbf{1}$ the all-ones vector, $L$ the sequence length, and $L'$ the block size.
\subsection{Autoregressive Language Modeling}
\label{sec:background-ar}
Autoregressive (AR) language models factorize the distribution over sequences $\x \in \mathcal{V}^L$ as $p_\theta(\x) = \prod_{\ell=1}^L p_\theta(\x^\supl \mid \x^{<\ell})$, where \mbox{$\x^{<\ell} = (\x^1, \ldots, \x^{\ell-1})$} denotes the prefix before position $\ell$. The Transformer architecture \citep{vaswani2017attentionneed} enables likelihood training in parallel, but generation is sequential and decodes a single token per forward pass.
\subsection{Discrete Diffusion Models}
\label{sec:background-discrete-diffusion}
Discrete diffusion models \citep{sohldickstein2015deepunsupervisedlearningusing, austin2023structureddenoisingdiffusionmodels, campbell2022continuoustimeframeworkdiscrete, lou2024discretediffusionmodelingestimating} define a family of increasingly noisy distributions $(q_t)_{t \in [0,1]}$ that interpolates from the data distribution $q_{\text{data}}$ at $t=0$ to a factorized noise distribution $\prod_{\ell=1}^L \mathrm{Cat}(\cdot; \prior)$ at $t=1$. Latent noisy sequences $\mathbf{z}_t \sim \prod_{\ell=1}^L q_t(\cdot | \x^\supl)$ are obtained via Markovian transitions, applied independently across positions. We focus on \emph{interpolating discrete diffusion processes}, whose \emph{forward process} $q_t(\cdot | \x^\supl)$ take the form:
\begin{equation}
\label{eq:forward-process}
\mathbf{z}_t^\supl \sim q_t(\cdot | \x^\supl; \alpha_t) = \mathrm{Cat}\big(\cdot\,; \alpha_t \x^\supl + (1 - \alpha_t) \prior\big),
\end{equation}
where $\alpha_t \in [0, 1]$ is a monotonically decreasing \emph{noise schedule} with $\alpha_0 \approx 1$ and $\alpha_1 \approx 0$. \Eqn{eq:forward-process}progressively corrupts $\x$ into a sample from the prior $\prior$. 
\paragraph{Generative Process}
To generate samples, diffusion models define a \emph{generative process} $p_\theta$ that reverses the forward process \Eqn{eq:forward-process}. Given a desired time-discretization \mbox{$0 = t_0 < t_1 < \cdots < t_{N_\text{step}} = 1$}, the generative process factors into the reverse trajectory as
{\small
\begin{equation}
p_\theta(\mathbf{z}_{t_0}, \ldots, \mathbf{z}_{t_{N_\text{step}}}) = p(\mathbf{z}_{t_{N_\text{step}}}) \prod_{i=1}^{N_\text{step}} p_\theta(\mathbf{z}_{t_{i-1}} \mid \mathbf{z}_{t_i}),
\end{equation}
\par}
where $p(\mathbf{z}_{t_{N_\text{step}}}) = \prod_{\ell} \mathrm{Cat}(\cdot; \prior)$ is the sequence-level prior. The generative transitions $p_\theta(\cdot \mid \mathbf{z}_t)$ have the same form as $q_{s|t}(\cdot \mid \mathbf{z}_t, \x)$ but replace $\x$ with a learned denoiser $\hat{\x}_\theta : \mathcal{V}^L \times [0,1] \to (\Delta^{\Vsize})^L$. We train the denoiser by minimizing the diffusion \emph{Negative Evidence Lower Bound} (NELBO) \citep{sohldickstein2015deepunsupervisedlearningusing, kingma2023variationaldiffusionmodels}. The form of the posterior $q_{s|t}$ and the NELBO depend on the choice of prior $\prior$.
\paragraph{Masked Diffusion Models (MDMs)}
\label{sec:background-mdm}
MDMs~\citep{sahoo2024simpleeffectivemaskeddiffusion, shi2025simplifiedgeneralizedmaskeddiffusion, ou2025absorbingdiscretediffusionsecretly} use a \emph{masked prior}, where $\prior = \mathbf{m} \in \mathcal{V}$ is the one-hot representation of a special token \mask. The forward process~\Eqn{eq:forward-process} preserves each token or replaces it with \mask. Once masked, a token remains in the absorbing state for the rest of the trajectory, and this carries over to the reverse process. The posterior distribution $q_{s|t}^{\text{MDM}}$ for $0 \leq s < t \leq 1$ follows from the Bayes rule.
\begin{equation}
\label{eq:mdm-posterior}
\begin{aligned}
q_{s|t}^{\text{MDM}}(\cdot | \mathbf{z}_t^\supl, \x^\supl) &= \begin{cases}
\mathrm{Cat}(\cdot\,; \boldsymbol{p}_{s|t}^\supl) & \text{if } \mathbf{z}_t^\supl = \mathbf{m}, \\[2pt]
\mathrm{Cat}(\cdot\,; \x^\supl) & \text{otherwise},
\end{cases} \\
\text{where}\quad \boldsymbol{p}_{s|t}^\supl &= \tfrac{\alpha_s - \alpha_t}{1 - \alpha_t} \x^\supl + \tfrac{1 - \alpha_s}{1 - \alpha_t} \mathbf{z}_t^\supl.
\end{aligned}
\end{equation}
A consequence of \Eqn{eq:mdm-posterior} is irreversibility. Once unmasked, a token cannot be revisited with ancestral sampling. Thus, mistakes are compounded. See \Eqn{eq:supp:mdm-nll} for the NELBO.
\paragraph{Uniform-State Diffusion Models (USDMs)}
\label{sec:background-usdm}
USDMs~\citep{schiff2025simpleguidancemechanismsdiscrete, sahoo2025diffusionduality} use a \emph{uniform prior} $\prior = \mathbf{1}/\Vsize$. Unlike MDMs, USDMs allow tokens to transition between any states throughout the generative process, enabling \emph{self-correction} of earlier mistakes. The posterior distribution $q_{s|t}^{\text{USDM}}$ takes the form
\begin{equation}
\label{eq:usdm-posterior}
q_{s|t}^{\text{USDM}}(\cdot | \mathbf{z}_t^\supl, \x^\supl) = \mathrm{Cat}\!\left(\cdot\,;\, \boldsymbol{\mu}_{s|t}^\supl / Z_{s|t}^\supl\right),
\end{equation}
where the (unnormalized) numerator $\boldsymbol{\mu}_{s|t}^\supl$ decomposes into four interpretable terms,
\begin{align}
\boldsymbol{\mu}_{s|t}^\supl ={}&
\Vsize \alpha_t (\mathbf{z}_t^\supl \!\odot\! \x^\supl)
+ \underbrace{(\alpha_{t|s} {-} \alpha_t)\mathbf{z}_t^\supl}_{\text{stay}} \nonumber \\
&+ \underbrace{(\alpha_s {-} \alpha_t)\x^\supl}_{\text{jump to } \x^\supl}
+ \underbrace{(1{-}\alpha_{t|s})(1{-}\alpha_s)\tfrac{\mathbf{1}}{\Vsize}}_{\text{uniform}},
\end{align}
with normalizer $Z_{s|t}^\supl = \Vsize \alpha_t \langle \mathbf{z}_t^\supl, \x^\supl \rangle + 1 - \alpha_t$ and $\alpha_{t|s} = \alpha_t / \alpha_s$. The three non-trivial components encode the three ways the next token is produced: \emph{stay} at the current token $\mathbf{z}_t^\supl$, \emph{jump} to the denoiser prediction $\x^\supl$, or \emph{sample uniformly} from the vocabulary. USDMs outperform MDMs in few-step generation~\citep{sahoo2025diffusionduality} and are better suited for guided generation~\citep{nisonoff2024unlocking, schiff2025simpleguidancemechanismsdiscrete, eyring2026ddno}. See \Eqn{eq:supp:duo-nll} for the NELBO.
\paragraph{Predictor-Corrector Samplers}
\label{sec:background-pc-samplers}
Predictor-Corrector (PC) samplers offer an alternative to ancestral sampling \citep{song2021scorebased, campbell2022continuoustimeframeworkdiscrete, gat2024discreteflowmatching, campbell2024generativeflowsdiscretestatespaces, wang2025remaskingdiscretediffusionmodels, deschenaux2026diffusiondualitychapterii}. They alternate or combine \emph{predictor} updates, which move from diffusion time $t$ to $s$ with $s < t$, with \emph{corrector} steps that inject noise. For MDMs, corrector steps re-mask tokens, letting the model revise earlier predictions. For USDMs, corrector steps re-inject random tokens, which can also improve quality, especially when scaling test-time compute \citep{deschenaux2026diffusiondualitychapterii}. We distinguish \emph{un-informed} correctors, which choose positions to re-noise uniformly at random, from \emph{informed} correctors, which use a per-token signal (entropy, likelihood, or a learned scorer) to target tokens most likely to be wrong \citep{zhao2025informedcorrectorsdiscretediffusion, liu2025thinkgeneratediscretediffusion, kim2025finetuningmaskeddiffusionprovable}. As an example, a simple \emph{un-informed} PC scheme first samples a clean proposal from the denoiser, then re-noises to time $s < t$ via the forward process \Eqn{eq:forward-process}:
\begin{align}
\tilde{\x}^\supl &\sim q_{0|t}(\cdot \mid \mathbf{z}_t^\supl,\hat{\x}_\theta^\supl(\mathbf{z}_t,t)), \\
\mathbf{z}_s^\supl &\sim q_s(\cdot \mid \tilde{\x}^\supl;\alpha_s) = \mathrm{Cat}\big(\cdot\,;\alpha_s \tilde{\x}^\supl + (1-\alpha_s)\prior\big).
\end{align}
\subsection{Block Diffusion Models}
\label{sec:background-block-diffusion}
Block diffusion models \citep{han2023ssdlmsemiautoregressivesimplexbaseddiffusion, arriola2025blockdiffusioninterpolatingautoregressive, wu2025fastdllmv2efficientblockdiffusion} combine an autoregressive factorization over blocks with masked diffusion within each block, generating tokens block-by-block from left to right. We partition a sequence $\x \in \mathcal{V}^L$ into $B = L/L'$ contiguous blocks $\x^1,\ldots,\x^B$ of fixed length $L' \leq L$, where $\x^b = [\x^{(b-1)L'+1}, \ldots, \x^{bL'}]$. The likelihood factorizes as
\begin{equation}
  \label{eq:block-diffusion-factorization}
  p_\theta(\x) = \prod_{b=1}^B p_\theta(\x^b \mid \x^{<b}),
\end{equation}
where each conditional $p_\theta(\x^b \mid \x^{<b})$ is parameterized with \emph{masked} diffusion (\sec{sec:background-discrete-diffusion}) and $\x^{<b}$ denotes the first $b-1$ blocks. Since \Eqn{eq:block-diffusion-factorization} is autoregressive over blocks, BDMs are often called semi-autoregressive (semi-AR). For block $b$, the forward process \Eqn{eq:forward-process} corrupts only tokens in $\x^b$, yielding noisy latents $\mathbf{z}_t^b$. The reverse transitions, conditioned on $\mathbf{z}_t^b$ and $\x^{<b}$, take the form
\begin{align}
\label{eq:block-reverse-process}
&p_\theta(\mathbf{z}_s^b \mid \mathbf{z}_t^b, \x^{<b}) \nonumber \\
&= \prod_{\ell=(b-1)L'+1}^{bL'} q_{s|t}(\cdot \mid \mathbf{z}_t^\supl, \hat{\x}_\theta^\supl(\mathbf{z}_t^b, \x^{<b}, t)).
\end{align}
Prior work \citep{arriola2025blockdiffusioninterpolatingautoregressive, wu2025fastdllmv2efficientblockdiffusion} instantiates $q_{s|t}$ using the masked-diffusion posterior \Eqn{eq:mdm-posterior}. Training minimizes the sum of per-block NELBOs.
\paragraph{KV Caching}
BDMs parameterize the denoiser $\hat{\x}_\theta$ with a Transformer that uses a \emph{block-causal} attention mask (see \supp{supp:attention-pattern} for details). Tokens in block $b$ attend bidirectionally within the block and causally to all tokens in the preceding blocks $\x^{<b}$. This pattern enables block-wise KV caching \citep{pope2022efficientlyscalingtransformerinference} at inference time. When generating block $b$, the keys and values for the already-generated prefix $\x^{<b}$ are reused across all diffusion steps within the block.
\section{BlockGen}
\label{sec:blockgen}
BlockGen is a blockwise sequence model trained over a mixture of block sizes, instantiated with masked or uniform diffusion within each block. \sec{sec:param-gen-model-blockgen} defines the mixture formulation and derives two likelihood bounds. \sec{sec:blockgen-gradient-estimation} presents the training objective and a stratified scheme to reduce variance. \sec{sec:hybrid-sampling-with-blockgen} introduces block-level predictor-correctors that score and re-noise unlikely tokens drawn from parallel decoding.
\subsection{Mixture Formulation over Block Sizes}
\label{sec:param-gen-model-blockgen}
We define BlockGen as a mixture over $M$ block-size-specific densities. Let $\mathcal{S} = \{s_1, \ldots, s_M\} \subset \mathbb{N}$ be a set of block sizes (e.g., $\{1, 2, 4, \ldots, 2^{M-1}\}$) and $\boldsymbol{\gamma} \in \Delta^M$ a mixture weight. The BlockGen density is defined as
\begin{equation}
\label{eq:blockgen-mixture}
  p_\theta^{\text{BlockGen}}(\x) = \sum_{i=1}^M \gamma_i \, p_\theta^{(s_i)}(\x),
\end{equation}
where each component $p_\theta^{(s_i)}$ is a valid density factorized into blocks of size $s_i$. BlockGen is agnostic to the paradigm used to model the block conditionals. We instantiate BlockGen with uniform and masked diffusion. Although uniform-state diffusion differs from masked diffusion only in the choice of prior $\boldsymbol{\pi}$, it improves the behavior in few steps (\sec{sec:experiments-math}) by allowing token revision, while masked diffusion remains competitive or stronger in likelihood. BlockGen supports both priors, and as we show in \sec{sec:experiments}, ARPC closes the gap between them and reverses the ranking at higher NFE. In the case $L' = 1$, BlockGen reduces to AR modeling.
\paragraph{Denoising backbone}
Rather than learning separate models for each $p_\theta^{(s_i)}$, we use a single shared denoiser $\hat{\x}_\theta(\cdot, \cdot, t; L')$ that takes the block size $L'$ as an additional input and sets the attention patterns accordingly. We follow the Diffusion Transformer (DiT) architecture \citep{arriola2025blockdiffusioninterpolatingautoregressive}, with one modification (\supp{supp:attention-pattern}). We optionally replace the block-causal attention with standard causal attention over the prefix, which enables sharing the KV cache across all block sizes during sampling.
\paragraph{Likelihood bounds}
We present two likelihood bounds that have different evaluation costs. The first is tighter, but requires computing $M$ ELBOs. The second is looser, but admits a cheap unbiased gradient estimator, and hence we use it for training (\sec{sec:blockgen-gradient-estimation}). Assume that each component $p_\theta^{(s_i)}$ in \Eqn{eq:blockgen-mixture} admits either a tractable likelihood or ELBO $\mathcal{E}^{(s_i)}(\theta, \x) \leq \log p_\theta^{(s_i)}(\x)$. When the likelihood is tractable, we set $\mathcal{E}^{(s_i)} = \log p_\theta^{(s_i)}$. The mixture density \Eqn{eq:blockgen-mixture} is then lower-bounded by
\begin{align}
\label{eq:blockgen-elbo}
  \log p_\theta^{\text{BlockGen}}(\x)
  &\;\geq\; \underbrace{\log \sum_{i=1}^{M} e^{\mathcal{E}^{(s_i)}(\theta, \x) + \log \gamma_i}}_{\text{log-sum-exp bound (eval)}} \nonumber \\
  &\;\stackrel{\text{Jensen}}{\geq}\; \underbrace{\sum_{i=1}^{M} \gamma_i \, \mathcal{E}^{(s_i)}(\theta, \x)}_\text{Mixture likelihood bound (train)}.
\end{align}
An alternative geometric-mean parameterization (\supp{app:geometric-mean-proof}) also admits the mixture likelihood bound as an ELBO. We do not pursue it here.
\subsection{Efficient Training}
\label{sec:blockgen-gradient-estimation}
We optimize $\theta$ by maximizing the ELBO. The gradient of the log-sum-exp bound takes the form
\begin{equation}
\label{eq:gbdm-gradient-lse}
\small
  \nabla_\theta \left[ -\log \sum_{i=1}^{M} \gamma_i \, e^{\mathcal{E}^{(s_i)}} \right]
  = -\sum_{i=1}^{M} \omega_i \, \nabla_\theta \mathcal{E}^{(s_i)}(\theta, \x),
\end{equation}
where $\omega_i = \gamma_i \, e^{\mathcal{E}^{(s_i)}(\theta, \x)} / \sum_{j=1}^{M} \gamma_j \, e^{\mathcal{E}^{(s_j)}(\theta, \x)}$.
Computing the gradient \Eqn{eq:gbdm-gradient-lse} requires evaluating the $M$ ELBOs, making training $M$ times more expensive than using a single block size. The mixture likelihood bound is cheaper, since it is an expectation over block sizes, $-\sum_{i=1}^{M} \gamma_i \mathcal{E}^{(s_i)}(\theta, \x) = \mathbb{E}_{i \sim \mathrm{Cat}(\boldsymbol{\gamma})}[-\mathcal{E}^{(s_i)}(\theta, \x)]$. A single sample $i \sim \mathrm{Cat}(\boldsymbol{\gamma})$ thus yields an unbiased gradient, so we optimize this looser bound during training.
\paragraph{Stratified block size selection}
\label{paragraph:stratified-block-size-selection}
When training on $D>1$ GPUs, we draw the per-GPU block sizes via stratified sampling rather than i.i.d. from $\boldsymbol{\gamma}$ or using the same block size on all GPUs, which reduces variance and improves the final model while preserving unbiasedness \citep{Kroese2011-xd}. We partition $(0,1]$ into $M$ bins via the cumulative sums of $\boldsymbol{\gamma}$, sample $u \sim \mathrm{Uniform}(0,1)$, and shift it by $D$ evenly-spaced offsets (wrapping modulo 1) to obtain one block size per GPU. See \algo{alg:stratified-block-size} for pseudocode.
\begin{tcolorbox}[
  colback=orange!12!white,
  colframe=orange!60!red!80!black,
  boxrule=1pt,
  arc=3pt,
  left=5pt, right=5pt,
  top=3pt, bottom=3pt,
  boxsep=2pt,
  before skip=4pt, after skip=4pt
]
\textbf{Training:} Optimize the mixture likelihood bound (cheap via stratified block selection). \\[4pt]
\textbf{Evaluation} (\supp{supp:valid-ppl})\textbf{:} Report the log-sum-exp bound by computing all $M$ component ELBOs.
\end{tcolorbox}
\subsection{Block-level Predictor-Correctors}
\label{sec:hybrid-sampling-with-blockgen}
Within a block, the denoiser produces factorized predictions over tokens, so independent sampling can yield combinations that are unlikely under the joint distribution \citep{xu2025energybaseddiffusionlanguagemodels}. We address this with predictor-corrector sampling that combines standard ancestral steps (\eqn{eq:mdm-posterior} for MDMs, \eqn{eq:usdm-posterior} for USDMs) with corrector steps \emph{informed} by the denoiser. An informed step scores each token of the proposal with a function $g$ and re-noises the lowest-scoring positions, presumed to be the worst tokens. We consider two strategies. The \emph{Entropy-Informed Predictor-Corrector} (EIPC) uses the per-token entropy of the diffusion-mode predictions and is applicable to any discrete diffusion model. The \emph{AR-Informed Predictor-Corrector} (ARPC) executes a second forward pass in AR mode ($L'{=}1$) and scores each token by the AR log-likelihood of the proposed value. The AR pass is enabled by the BlockGen mixture (\sec{sec:param-gen-model-blockgen}), which trains the same denoiser at $L'{=}1$ and at the larger block size used for sampling. EIPC is the natural baseline for ARPC: both are informed correctors, but we do not need to train over multiple block sizes to implement EIPC. Since ARPC generally outperforms EIPC (\sec{sec:experiments-math}), this justifies training over a mixture of block sizes.
\paragraph{Informed steps}
At diffusion time $t$, let $p^{\mathrm{D}} = \hat{\x}_\theta(\mathbf{z}_t^b, \x^{<b}, t; L')$ be the output of the denoiser. A standard ancestral sampling step draws $\mathbf{z}_s^b \sim q_{s|t}(\cdot \mid \mathbf{z}_t^b, p^{\mathrm{D}})$. Informed steps instead sample a clean proposal $\tilde{\x}^b \sim q_{0|t}(\cdot \mid \mathbf{z}_t^b, p^{\mathrm{D}})$, score each position with $g$, and re-noise the lowest-scoring $k = \texttt{round}((1{-}\alpha_s) L')$ positions to the noise level at time $s$. EIPC uses $g^\ell = \sum_v p^{\mathrm{D},\ell}(v) \log p^{\mathrm{D},\ell}(v)$, the negative entropy of $p^{\mathrm{D},\ell}$. ARPC executes a second forward pass at $L'=1$ to obtain $p^{\mathrm{AR}} = \hat{\x}_\theta(\tilde{\x}^b, \x^{<b}; L'{=}1)$ and sets $g^\ell = \log p^{\mathrm{AR},\ell}(\tilde{\x}^{b,\ell})$, the AR log-likelihood of the proposed token. Thus, ARPC does not require a separate verifier. At matched NFE, ARPC outperforms EIPC (\sec{sec:experiments-math}).
\paragraph{Step schedule}
We use informed steps after $n_\text{warmup}$ ancestral diffusion steps and every $\text{GE} \geq 1$ updates afterwards. \Cref{alg:EIPC,alg:ARPC} presents the two samplers, with \textsc{Informed}$(i)$ true when step $i$ is an informed step.
\section{Experiments}
\label{sec:experiments}
Most prior work measures the sample quality via the Generative Perplexity (Gen. PPL, \supp{app:gen-ppl}) because small models fail on most open-ended benchmarks. Low perplexity, however, does not imply correctness or high quality \citep{velickovic2026perplexitycannotalwaystellright, feng2025theoreticalbenefitlimitationdiffusion, deschenaux2024promisesoutlookschallengesdiffusion}. We therefore evaluate primarily on GSM8K \citep{cobbe2021trainingverifierssolvemath}, where every problem comes with a ground-truth solution. \sec{sec:experiments-math} reports the accuracy on GSM8K. \sec{sec:experiments-density-estimation} reports the validation likelihood and the Generative Perplexity / entropy frontier on OpenWebText \citep{Gokaslan2019OpenWeb}.
\subsection{Mathematical Reasoning}
\label{sec:experiments-math}
\begin{figure*}[t]
  \centering
  \begin{subfigure}[t]{0.49\textwidth}
    \centering
    \includegraphics[width=\textwidth]{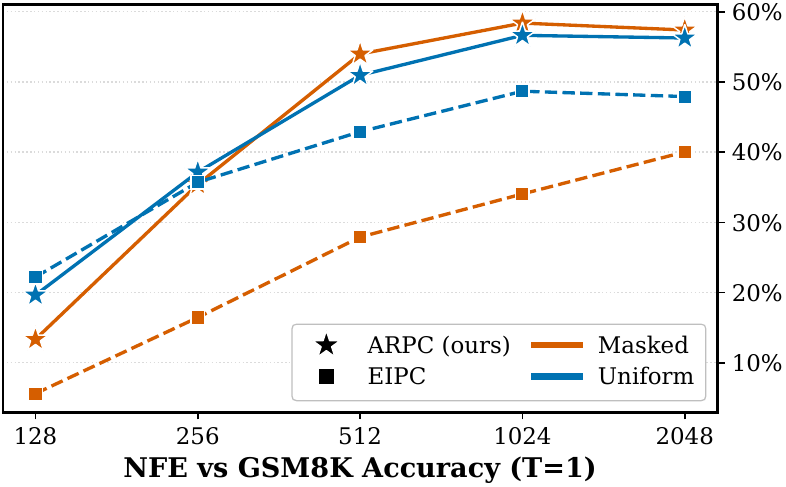}
    \label{fig:arpc_vs_eipc_b16_T1}
  \end{subfigure}\hfill
  \begin{subfigure}[t]{0.49\textwidth}
    \centering
    \includegraphics[width=\textwidth]{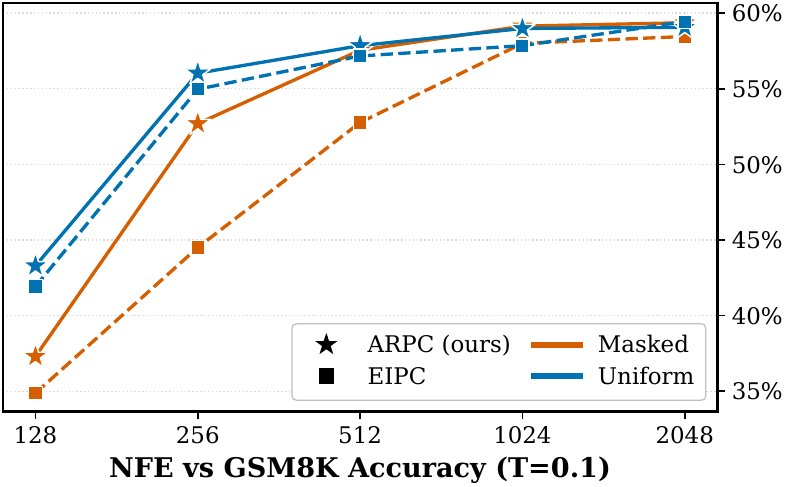}
    \label{fig:arpc_vs_eipc_b16_T0p1}
  \end{subfigure}
  \caption{\textbf{ARPC vs EIPC GSM8K accuracy with block size 16} as a function of NFE. Models are trained on TinyGSM and evaluated on the GSM8K test set. Each curve shows the best performance for a given NFE. \emph{Entropy-Informed Predictor-Corrector} (EIPC) uses a single-block-16 model, and ARPC uses the multi-block mixture from \Eqn{eq:blockgen-mixture} with $\gamma_1 = 0.05$, $\gamma_{16} = 0.95$. ARPC outperforms EIPC across most NFE budgets (left and right).}
  \label{fig:arpc_vs_eipc_b16}
\end{figure*}
\paragraph{Experimental Setup}
TinyGSM \citep{liu2023tinygsm} contains $\sim$11.8M GSM8K-style synthetic word problems \citep{cobbe2021trainingverifierssolvemath} with associated Python programs. \textbf{We train on TinyGSM and report exact-match accuracy on the GSM8K test set} after executing the generated code. We train a 170M-parameter modified DiT for 250k steps; tokenizer, architecture, and optimization details are deferred to \supp{sec:appendix-experimental-details}. We train BlockGen with block sizes $\{1, L'\}$ for $L' \in \{16, 32\}$ to enable ARPC, setting small $\gamma_1 \in \{0.05, 0.10, 0.15\}$ with $\gamma_{L'} = 1 - \gamma_1$, since the AR pass is used only as a verifier. We compare BlockGen with AR, MDLM \citep{sahoo2024simpleeffectivemaskeddiffusion}, Duo \citep{sahoo2025diffusionduality}, and single-block-size BDM \citep{arriola2025blockdiffusioninterpolatingautoregressive}.
\paragraph{Ancestral vs ARPC}
\Fig{fig:ar_vs_ancestral_vs_arpc_b16} shows that under ancestral sampling, uniform diffusion outperforms masked in the block-by-block setting, as for full-sequence diffusion \citep{deschenaux2026diffusiondualitychapterii}. However under ARPC the gap shrinks. Uniform is best at low NFE, masked at higher NFE, but the two stay within a small margin of each other. The same pattern appears in Generative Perplexity on OpenWebText (\sec{sec:experiments-density-estimation}).
\paragraph{ARPC outperforms ancestral and EIPC}
EIPC applies to any single-block model, while ARPC requires the BlockGen mixture to enable the AR pass; comparing the two isolates whether the AR-based score is worth training on multiple block sizes. At $T{=}1$, ARPC reaches higher accuracy than ancestral and EIPC across the NFE range, with the largest gap on masked diffusion, and ARPC (uniform) matches AR with sampling at $256$ NFE (\fig{fig:ar_vs_ancestral_vs_arpc_b16},~\fig{fig:arpc_vs_eipc_b16}). At $T{=}0.1$, EIPC approaches ARPC and remains slightly behind. AR with greedy decoding remains the strongest baseline.
\paragraph{Impact of the Block size}
\fig{fig:arpc_b16_vs_b32} compares ARPC at block sizes $16$ and $32$ at matched NFEs. Models with a block size of $16$ generally reach a higher accuracy. Recall that the ELBO loosens as the block size grows \citep{arriola2025blockdiffusioninterpolatingautoregressive}, so this ordering is expected. Block size $32$, mixture-weight $\bm{\gamma}$ ablations, temperature sweeps, and raw accuracy tables are in \supp{app:raw-accuracies} and \supp{app:sampler-ablations}.
\subsection{Language Modeling on OpenWebText and LM1B}
\label{sec:experiments-density-estimation}
We evaluate BlockGen on LM1B \citep{chelba2014billionwordbenchmarkmeasuring} and OpenWebText (OWT) \citep{Gokaslan2019OpenWeb}, reporting validation perplexity of BlockGen across mixture weights and comparing sample quality between AR, ancestral sampling, EIPC, and ARPC.

\paragraph{Experimental Setup}
We train a 170M-parameter modified DiT for 1M steps on OpenWebText (GPT-2 tokenizer, context length $1024$) using the mixture likelihood bound \Eqn{eq:blockgen-elbo}; tokenizer, architecture, and optimization details are in \supp{sec:appendix-experimental-details}. We report validation perplexity (Val.\ PPL) under the log-sum-exp bound (\supp{supp:mc-elbo-blockgen}) and Generative Perplexity (Gen.\ PPL) under GPT-2 Large \citep{Radford2019LanguageMA}, with per-sample unigram entropy $H_1$ as a diversity proxy. Baselines: AR, single-block-size BDM \citep{arriola2025blockdiffusioninterpolatingautoregressive}, MDLM \citep{sahoo2024simpleeffectivemaskeddiffusion}, Duo \citep{sahoo2025diffusionduality}, SEDD \citep{lou2024discretediffusionmodelingestimating}, and UDLM \citep{schiff2025simpleguidancemechanismsdiscrete}.
\paragraph{Single-block baselines}
For Masked BDMs, BD3-LM \citep{arriola2025blockdiffusioninterpolatingautoregressive} uses a costly variance-reduction scheme that optimizes the noise schedule every 5k steps over the validation set. We find that training with unweighted cross-entropy matches their performance on OWT and slightly improves it on LM1B, without the overhead (\tab{tab:valid-ppl-after-250k-steps}, \supp{app:owt-val-ppl}). Recent work shows that unweighted CE optimizes the true NELBO \citep{shi2025demystifyingdiffusionobjectivesreweighted,sahoo2026scalingbeyondmasked,sahoo2025esotericlanguagemodels}. For uniform-state BDMs, unweighted CE underperforms the NELBO on OWT for block sizes $>4$, so we use the NELBO for all $L'>1$ \citep{vonrütte2025generalizedinterpolatingdiscretediffusion}.
\paragraph{Tight likelihood interpolation}
\begin{table}[t]
  \caption{Validation perplexity (Val. PPL) on OWT after 1M training steps. \textbf{Takeaway:} BlockGen closes the gap between block diffusion and autoregressive models, achieving 17.5 PPL vs.\ 16.7 for AR. ${}^\dagger$Adapted from MDLM checkpoint (850k steps). ${}^\ddagger$From \citep{sahoo2025diffusionduality}. \textbf{Bold}/\underline{underline}: best/second-best block diffusion. Training Masked BDMs with unweighted CE improves likelihood.}
  \label{tab:owt-paradigm-comparison}
  \centering
  \footnotesize
  \setlength{\tabcolsep}{6pt}
  \renewcommand{\arraystretch}{1.1}
  {%
  \newcommand{\tabrow}{\hspace*{0.6em}}
\begin{tabularx}{\columnwidth}{@{} >{\raggedright\arraybackslash}X c@{}}
\toprule
Model & Val.\ PPL \\
\midrule
\multicolumn{2}{@{}l@{}}{\textit{Autoregressive}} \\
\tabrow Transformer & 16.7 \\
\midrule
\multicolumn{2}{@{}l@{}}{\textit{Sequence Diffusion (Masked)}} \\
\tabrow SEDD Absorb${}^\ddagger$ \citep{lou2024discretediffusionmodelingestimating} & 24.1 \\
\tabrow MDLM${}^\ddagger$ \citep{sahoo2024simpleeffectivemaskeddiffusion} & 23.2 \\
\midrule
\multicolumn{2}{@{}l@{}}{\textit{Sequence Diffusion (Uniform)}} \\
\tabrow SEDD Uniform${}^\ddagger$ \citep{lou2024discretediffusionmodelingestimating} & 29.7 \\
\tabrow UDLM${}^\ddagger$ \citep{schiff2025simpleguidancemechanismsdiscrete} & 27.4 \\
\tabrow Duo${}^\ddagger$ \citep{sahoo2025diffusionduality} & 25.2 \\
\midrule
\multicolumn{2}{@{}l@{}}{\textit{BDM (single-block, $L' = 16$)}} \\
\tabrow BDM${}^\dagger$ \citep{arriola2025blockdiffusioninterpolatingautoregressive} & 22.3 \\
\tabrow Masked (CE) & 21.6 \\
\tabrow Uniform (ELBO) & 23.6 \\
\midrule
\multicolumn{2}{@{}p{\columnwidth}@{}}{\cellcolor{gray!15}\textit{BlockGen (ours, $L_{\max}{=}16$, $\gamma_1{=}0.05$, $\gamma_{16}{=}0.95$)}} \\
\cellcolor{gray!15}\tabrow Masked & \cellcolor{gray!15}19.1 \\
\cellcolor{gray!15}\tabrow Uniform & \cellcolor{gray!15}20.9 \\
\midrule
\multicolumn{2}{@{}p{\columnwidth}@{}}{\cellcolor{gray!15}\textit{BlockGen (ours, $L_{\max}{=}16$, $\gamma_i{=}0.2$)}} \\
\cellcolor{gray!15}\tabrow Masked & \cellcolor{gray!15}\textbf{17.5} \\
\cellcolor{gray!15}\tabrow Uniform & \cellcolor{gray!15}\underline{18.5} \\
\bottomrule
\end{tabularx}}
\end{table}

\tab{tab:owt-paradigm-comparison} reports the 1M-step Val. PPL on OWT under two $\bm{\gamma}$ regimes at $L_{\max}{=}16$ (block sizes $\{1, 2, 4, 8, 16\}$). With uniform weights $\gamma_i{=}0.2$, BlockGen reaches $\mathbf{17.5}$ PPL with masked diffusion ($0.8$ below AR at $16.7$, $4.1$ below the best single-block BDM at $21.6$) and $18.5$ with uniform diffusion. With $\gamma_1{=}0.05$, $\gamma_{16}{=}0.95$, close to single-block-$16$ training, BlockGen still reaches $19.1$ (masked) and $20.9$ (uniform), $2$--$3$ PPL below the single-block BDMs ($21.6$ and $23.6$). We use stratified block-size selection (\sec{paragraph:stratified-block-size-selection}) for all 1M runs.
\paragraph{Evaluating the sample quality}
We sweep $T \in \{0.50, 0.55, \ldots, 1.20\}$ and plot the Gen.\ PPL / entropy frontier \citep{pynadath2025candihybriddiscretecontinuousdiffusion} across AR, single-block ancestral, EIPC, and ARPC; further details are in \supp{app:owt-additional-figures}.
\begin{figure*}[t]
  \centering
  \begin{subfigure}[t]{0.32\textwidth}
    \centering
    \includegraphics[width=\textwidth]{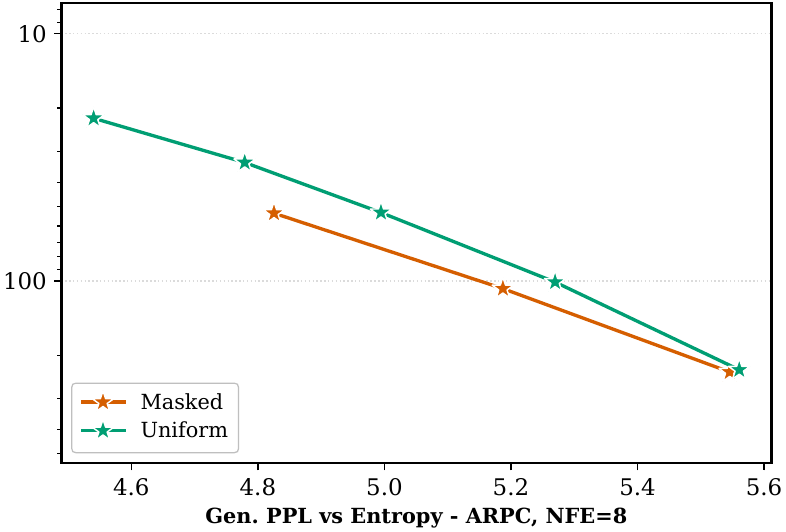}
    \label{fig:owt-arpc-mvu-nfe8}
  \end{subfigure}\hfill
  \begin{subfigure}[t]{0.32\textwidth}
    \centering
    \includegraphics[width=\textwidth]{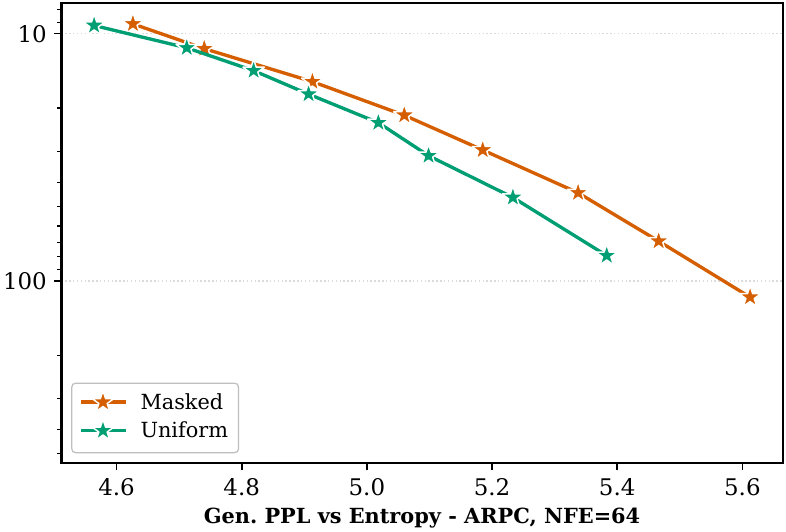}
    \label{fig:owt-arpc-mvu-nfe64}
  \end{subfigure}\hfill
  \begin{subfigure}[t]{0.32\textwidth}
    \centering
    \includegraphics[width=\textwidth]{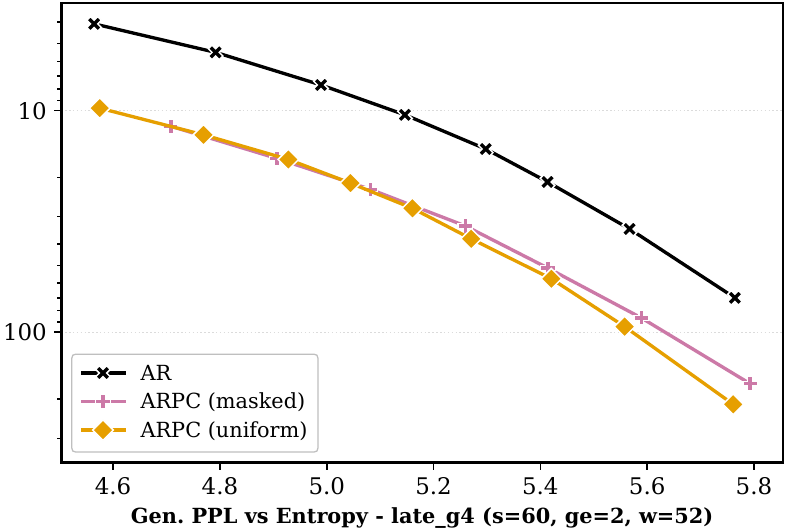}
    \label{fig:owt-arpc-vs-ar-nfe64}
  \end{subfigure}
  \caption{\textbf{Sample quality on OpenWebText.} Each curve represents a sweep over temperatures with fixed NFEs. Lower Gen. PPL at matched entropy is better. \textbf{Left:} masked vs uniform diffusion under ARPC at per-block NFE $=8$. Uniform-ARPC reaches better Gen. PPL than masked-ARPC. \textbf{Middle:} same comparison at per-block NFE $=64$. Masked-ARPC has the better frontier. \textbf{Right:} ARPC vs AR at per-block NFE $=64$ with a late-correction schedule ($60$ predictor steps, one corrector every $2$ steps after a $52$-step warmup). AR keeps lower Gen. PPL across the practical temperature range, with masked-ARPC closer to AR than uniform-ARPC. Three other late-correction schedules give qualitatively similar curves (\supp{app:owt-additional-figures}).}
  \label{fig:owt-arpc-mvu}
\end{figure*}
\paragraph{Behavior of Masked vs Uniform on OWT}
With the (single block) ancestral sampler, uniform diffusion reaches lower Gen. PPL than masked across the NFE budgets we tested, with the gap narrowing at higher NFE (\supp{app:owt-additional-figures}). With ARPC, at $8$ NFE per block, uniform-ARPC has the stronger frontier, while at $64$ NFE per block masked-ARPC performs better (\fig{fig:owt-arpc-mvu}, left and middle). This matches the qualitative GSM8K trend in \sec{sec:experiments-math}.
\paragraph{ARPC narrows but does not close the gap to AR}
\fig{fig:owt-arpc-mvu} (right) compares ARPC with NFE $=64$ per block against AR. AR reaches lower Gen. PPL across the practical temperature range, and masked-ARPC is closer to AR than uniform-ARPC, consistent with the middle panel. Three other schedules at NFE $=64$ give qualitatively similar curves (\supp{app:owt-additional-figures}).
\paragraph{Matched total-NFE comparison}
Prior work observed that block-by-block masked diffusion reaches higher Gen.\ PPL than full-sequence masked diffusion at matched total NFE \citep{sahoo2025esotericlanguagemodels}. We extend this observation to uniform diffusion: at NFE $=1024$ on OWT, MDLM and Duo reach lower Gen.\ PPL than block-by-block ARPC under either prior (\fig{fig:owt-frontier-matched}, \supp{app:owt-additional-figures}). However, BlockGen is faster than full-sequence diffusion during sampling because it caches the KV of decoded blocks.
\section{Related Work}
\paragraph{Masked vs uniform discrete diffusion}
The case for USDMs draws on three recent observations: equal or higher downstream accuracy despite worse perplexity \citep{vonrutte2025scalingbehaviordiscretediffusion, sahoo2026scalingbeyondmasked}, stronger test-time scaling under predictor-corrector sampling \citep{deschenaux2026diffusiondualitychapterii}, and better scaling in the data-constrained regime \citep{vonrutte2025scalingbehaviordiscretediffusion}. USDMs also implement guidance more naturally than MDMs \citep{schiff2025simpleguidancemechanismsdiscrete, eyring2026ddno}. Prior work attributes most test-time advantages to self-correction, since tokens can be re-sampled at each step, while masked tokens are fixed once unmasked \citep{vonrutte2025scalingbehaviordiscretediffusion}. This holds when correctors re-noise tokens at random. However, with informed correctors \citep{zhao2025informedcorrectorsdiscretediffusion, liu2025thinkgeneratediscretediffusion}, MDMs can also correct earlier mistakes.
\paragraph{Predictor-corrector samplers}
With ancestral sampling, MDMs fix each token once it is unmasked, so parallel-decoding errors cannot be revised later \citep{wang2025remaskingdiscretediffusionmodels}. Predictor-corrector samplers address this by re-masking; re-injecting noise also helps USDMs \citep{deschenaux2026diffusiondualitychapterii}. Random correctors re-noise at uniformly chosen positions \citep{campbell2022continuoustimeframeworkdiscrete, campbell2024generativeflowsdiscretestatespaces, gat2024discreteflowmatching, wang2025remaskingdiscretediffusionmodels, deschenaux2026diffusiondualitychapterii}. \emph{Informed} correctors train an additional model to choose which positions to re-sample \citep{lezama2023discrete, liu2025thinkgeneratediscretediffusion, zhao2025informedcorrectorsdiscretediffusion, kim2025finetuningmaskeddiffusionprovable, zhang2025correctivediffusionlanguagemodels, peng2025pathplanningmaskeddiffusion, peng2025plannerawarepathlearning}; EIPC and ARPC instead reuse the base model directly. Adaptive parallel decoding (APD) \citep{israel2025adaptiveparalleldecoding} mixes AR and diffusion likelihoods using two separate models to decide which tokens to accept; ARPC instead uses a single shared denoiser, with AR predictions only deciding which tokens to re-generate. APD only considers masked diffusion, while we also study USDMs.
%
\paragraph{Hybrid AR-diffusion approaches}
Hybrid AR-diffusion methods combine AR and discrete diffusion to retain KV caching during parallel decoding. SSD-LM \citep{han2023ssdlmsemiautoregressivesimplexbaseddiffusion} uses block-wise generation on a continuous probability simplex, and BD3-LM \citep{arriola2025blockdiffusioninterpolatingautoregressive} introduced discrete single-block-size diffusion. TiDAR \citep{liu2025tidarthinkdiffusiontalk} adapts an AR checkpoint to a masked multi-token generator with AR verification, and FastDLM \citep{wu2025fastdllmv2efficientblockdiffusion, wang2025diffusionllmsfasterthanarinference} adapts MDMs for block-wise caching. CtrlDiff \citep{huang2025ctrldiffboostinglargediffusion} chooses block sizes in inference via heuristics or RL. Eso-LMs \citep{sahoo2025esotericlanguagemodels} train a mixture of MDM and AR objectives. Their sampling generates a draft with diffusion and then completes the rest autoregressively, bringing them closer to any-order AR models \citep{pmlr-v32-uria14, strauss2021arbitraryconditionaldistributionsenergy, hoogeboom2022autoregressivediffusionmodels, shih2022traininginferenceanyorderautoregressive, kim2025anyorderflexiblelengthmasked}. BlockGen trains a single model over a mixture of block sizes and remains agnostic to the paradigm within each block. Thus BlockGen can freely interleave AR and diffusion predictions.
\section{Conclusion}
\label{sec:conclusion}
We presented BlockGen, a framework for blockwise sequence modeling that trains a single denoiser over a mixture of block sizes. Training over multiple block sizes lets the same denoiser act as a block-by-block sampler and as an AR verifier, without separate models. Empirically, BlockGen narrows the OpenWebText likelihood gap to autoregressive models ($17.5$ PPL with masked diffusion and uniform $\bm{\gamma}$ vs.\ $16.7$ for AR, against $21.6$ for the best fixed-block BDM), and \emph{ARPC} (AR-informed predictor-corrector sampling) improves GSM8K accuracy at matched NFEs (with models trained on TinyGSM). The same setup lets us revisit the question \emph{is uniform the stronger paradigm for discrete diffusion?} Under ancestral sampling we recover the prior finding that uniform outperforms masked, but under ARPC the gap closes and the ranking reverses at higher NFE on both GSM8K and OpenWebText.

\section*{Limitations}
\label{sec:limitations}
The main limitation of this work is scale. We train BlockGen on LM1B and OpenWebText with a $170$M-parameter backbone, following the settings of Block Diffusion \citep{arriola2025blockdiffusioninterpolatingautoregressive}, and do not establish whether the same tradeoffs persist for larger-scale LLMs.

Training block sequence models such as Block Diffusion or BlockGen is more costly than AR and full-sequence discrete diffusion models, because the network processes a doubled sequence length (\sec{supp:attention-pattern}) to attend to both the clean and the noisy sequence. We do not address this overhead and leave it to future work.

Beyond scale and training cost, our empirical claims are scoped to specific samplers and budgets. On OpenWebText, ARPC does not surpass AR in Generative Perplexity in the practical low-temperature regime under any of the corrector schedules we tested at $64$ NFE per block. The relative ordering between ARPC and EIPC depends on the prior, the temperature, and the NFE: on uniform diffusion, EIPC matches or slightly surpasses ARPC at certain operating points, consistent with the low-temperature trend on GSM8K. We compare ARPC against AR with sampling; greedy AR remains the strongest baseline on GSM8K at all NFE budgets we considered. The masked-vs-uniform regime shift we report is therefore a regime-dependent observation rather than a universal ranking.

\newpage

\medskip

\bibliography{references}

\newpage


\appendix

\section{Ethical considerations}
\label{sec:impact-statement}
This paper presents work whose goal is to advance the field of Machine Learning, specifically in the area of language modeling. As with any language modeling research, there are inherent risks, including the potential for generating misleading information, fake news, or harmful content. However, at the current scale of our models, these concerns are minimal, as the capabilities of our approach remain substantially below those of state-of-the-art autoregressive large language models. The primary contribution of this work is methodological.

\paragraph{Datasets and licenses}
We use four publicly available datasets, all consistent with their intended use for language-modeling research:
\textbf{GSM8K} \citep{cobbe2021trainingverifierssolvemath} is released under the MIT license and contains $\sim$8.5K grade-school math word problems with step-by-step solutions.
\textbf{TinyGSM} \citep{liu2023tinygsm} is released under the MIT license and contains $\sim$11.8M GSM8K-style synthetic word problems with Python programs; it was generated with GPT-3.5.
\textbf{One Billion Words (LM1B)} \citep{chelba2014billionwordbenchmarkmeasuring} is a standard English language-modeling benchmark released for academic research; the original distribution at \url{statmt.org/lm-benchmark} does not specify a license, and the dataset is widely re-distributed (e.g., via TFDS under Apache 2.0).
\textbf{OpenWebText} \citep{Gokaslan2019OpenWeb} is released under CC0 (public domain) and is an open replication of the WebText corpus crawled from Reddit-linked pages; users should be aware that the underlying web data may contain noise, biases, and occasional offensive language inherent to web text. We use these datasets in their published, anonymized forms; none of them contain personally identifying information about private individuals to the best of our knowledge.

\section{Additional Experimental Details}
\label{sec:appendix-experimental-details}
\subsection{Data Preparation}
\begin{figure*}[t]
  \centering
  \begin{subfigure}[t]{\textwidth}
      \centering
      \includegraphics[width=\textwidth]{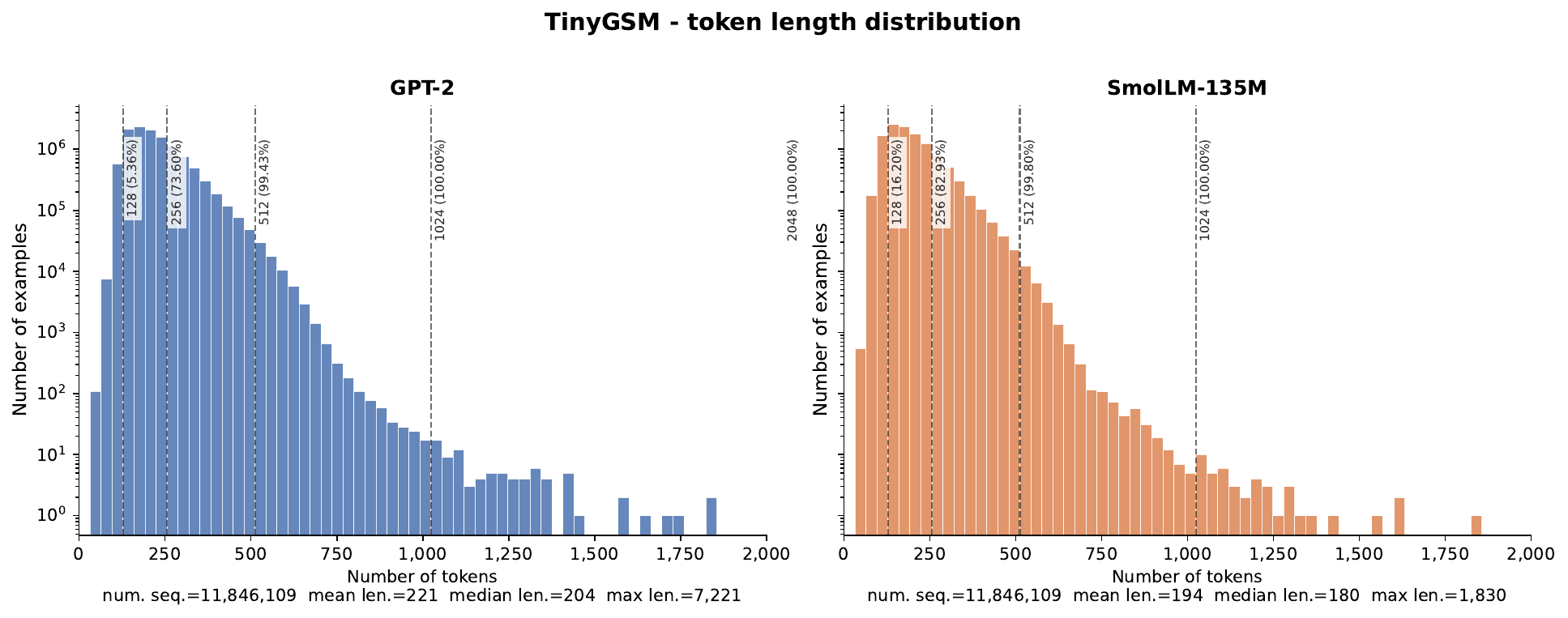}
      \caption{GPT-2 vs.\ \texttt{SmolLM-135M}.}
      \label{fig:tinygsm-gpt2-smollm}
  \end{subfigure}

  \vspace{0.5em}
  \begin{subfigure}[t]{\textwidth}
      \centering
      \includegraphics[width=\textwidth]{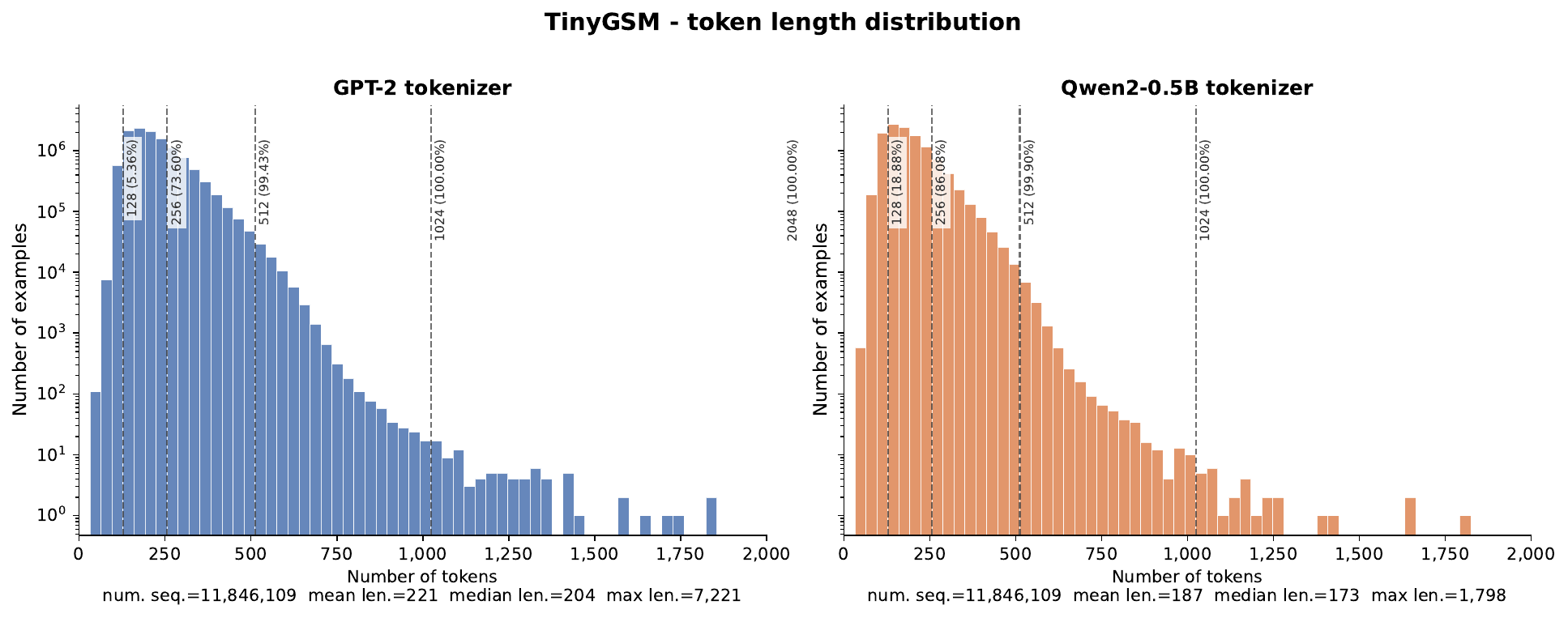}
      \caption{GPT-2 vs.\ Qwen2.5.}
      \label{fig:tinygsm-gpt2-qwen2}
  \end{subfigure}
  \caption{\textbf{Distribution of tokenized sequence lengths on TinyGSM.} The GPT-2 tokenizer was trained on web text and produces longer tokenized sequences on code than tokenizers trained on code. We compare GPT-2 against \texttt{SmolLM-135M} (top) and Qwen2.5 (bottom). Both tokenizers are trained on code, and produce shorter sequences.}
  \label{fig:tinygsm-tokenizer-compare}
\end{figure*}

For the One Billion Words (LM1B) dataset, we follow the preprocessing in prior work \citep{lou2024discretediffusionmodelingestimating, sahoo2024simpleeffectivemaskeddiffusion}\footnote{\url{https://github.com/louaaron/Score-Entropy-Discrete-Diffusion/blob/main/data.py}} and tokenize with \texttt{bert-base-uncased} following DiffusionBERT \citep{he2022diffusionbertimprovinggenerativemasked}. We use both wrapped and non-wrapped variants \citep{sahoo2025diffusionduality, arriola2025blockdiffusioninterpolatingautoregressive}. For OpenWebText, we tokenize with \texttt{GPT2}, concatenate sequences to length 1024 with \texttt{eos} tokens between them, and reserve the last 100k documents for validation.
\paragraph{TinyGSM tokenization}
\label{app:tinygsm-tokenizer}
\fig{fig:tinygsm-tokenizer-compare} shows that tokenizers trained on code (such as \texttt{SmolLM-135M} and Qwen2.5 \citep{qwen2025qwen25technicalreport}) compress TinyGSM examples to noticeably shorter sequences than GPT-2 \citep{Radford2019LanguageMA}. Qwen2.5 yields even higher compression than \texttt{SmolLM}, but its vocabulary contains $\sim$151k tokens against $\sim$50k for both \texttt{SmolLM} and GPT-2. The larger vocabulary increases the embedding and output projection parameter counts, so we use \texttt{SmolLM}.
\subsection{Denoising Backbone}
We parameterize all models using the modified diffusion transformer architecture \citep{peebles2023scalablediffusionmodelstransformers} from recent discrete diffusion papers \citep{lou2024discretediffusionmodelingestimating, sahoo2024simpleeffectivemaskeddiffusion}. We use 12 layers, a hidden dimension of 768, 12 attention heads, and a timestep embedding dimension of 128. Following prior work \citep{sahoo2024simpleeffectivemaskeddiffusion, schiff2025simpleguidancemechanismsdiscrete, arriola2025blockdiffusioninterpolatingautoregressive, sahoo2025diffusionduality}, we keep the adaptive layernorm but feed it a zero vector, making the denoiser time-unconditional. This contrasts with prior work on USDMs \citep{schiff2025simpleguidancemechanismsdiscrete, sahoo2025diffusionduality}, where time-conditioning is used to improve performance. We chose this approach for simplicity, to avoid modifying the backbone. In principle, one could enable both time-conditioning and KV-caching by using a zero time embedding for clean tokens (so their activations remain cacheable during sampling) and using the actual time embedding for noisy blocks only. However, this introduces differences in the denoiser compared to prior work, therefore, we keep the denoiser time-unconditional in this work. If anything, making the denoiser time-conditional should improve the performance compared to this work.
\subsection{Attention Implementation}
\label{supp:attention-pattern}
\begin{figure*}[t]
  \centering
  \begin{subfigure}[t]{0.48\textwidth}
      \centering
      \includegraphics[width=\textwidth]{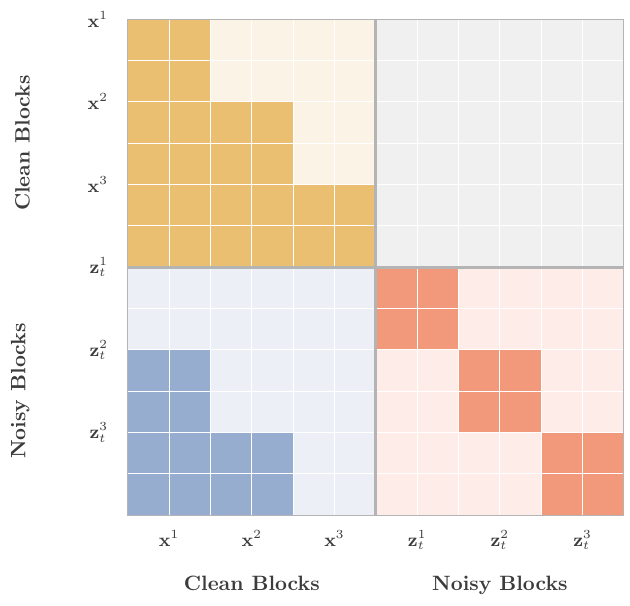}
      \caption{Standard block diffusion attention.}
      \label{fig:attn-standard}
  \end{subfigure}
  \hfill
  \begin{subfigure}[t]{0.48\textwidth}
      \centering
      \includegraphics[width=\textwidth]{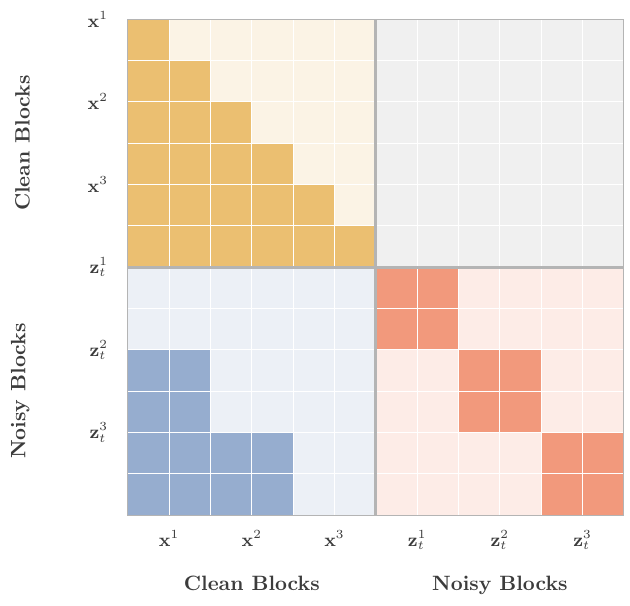}
      \caption{Block Diffusion with Causal attention on the clean prefix.}
      \label{fig:attn-causal}
  \end{subfigure}
  \caption{\textbf{Attention patterns for block diffusion (training)} with $L' = 2$. Noisy blocks attend to all tokens within the block and to all clean tokens in previous blocks. (a) shows the attention from BD3-LM \citep{arriola2025blockdiffusioninterpolatingautoregressive}. (b) uses causal attention over the clean tokens. This is important for BlockGen since we train over multiple block sizes but want to share a single KV cache across all block sizes during inference. Unlike BD3-LM \citep{arriola2025blockdiffusioninterpolatingautoregressive}, we concatenate the clean context before the noisy tokens. This preserves correctness and does not affect performance, with a more intuitive pattern where tokens attend only to previous tokens in the input.}
  \label{fig:attention-patterns}
\end{figure*}

\Fig{fig:attention-patterns} shows the attention pattern used during training. Following BD3-LM \citep{arriola2025blockdiffusioninterpolatingautoregressive}, we feed the denoising network sequences of 2x the context length (256 on LM1B, 2048 on OWT) to obtain predictions for all noisy blocks conditioned on the clean prefix. We use a slightly modified version of their FlexAttention \citep{dong2024flexattentionprogrammingmodel} attention masks. While BD3-LM \citep{arriola2025blockdiffusioninterpolatingautoregressive} concatenates sequences with clean tokens second, we place the clean context first and noisy tokens second. This ordering does not affect speed or correctness but yields a more natural attention pattern that mirrors standard AR models, where tokens attend to the past rather than the future.
\subsection{Optimization hyperparameters}
We use the Adam optimizer with $\beta_1 = 0.9$, $\beta_2 = 0.999$, batch size 512, learning rate $3e-4$ with linear warmup over 2500 steps, no cooldown, and no weight decay. We train in mixed precision with gradient clipping to 1 for 1M steps. We maintain an exponential moving average of the weights with decay 0.9999 and use a dropout rate of 0.1. We do not tune hyperparameters further, following prior work and to save compute.
\subsection{Training Cost}
\label{app:training-cost}
\tab{tab:training-cost} shows the training cost for all models. Training BlockGen over a mixture of block sizes does not change the per-step cost compared to a single block size. We use FlexAttention \citep{dong2024flexattentionprogrammingmodel} kernels that are compiled once per block size and reused across steps. With $M$ block sizes we cache $M$ kernels. We use at most $M=6$ in our experiments.
\subsection{Evaluating the Sample Quality}
\label{app:eval-sample-quality}
\paragraph{Generative Perplexity (Gen.\ PPL).}
\label{app:gen-ppl}
We evaluate the quality of generated text using the perplexity under a larger reference language model (GPT-2 Large), following prior work \citep{lou2024discretediffusionmodelingestimating, sahoo2024simpleeffectivemaskeddiffusion, sahoo2025diffusionduality}. Given $N$ generated sequences $\{\x^{(i)}\}_{i=1}^N$ (each tokenized with the GPT-2 tokenizer and of length $L$), we compute
{\small
\begin{align}
&\text{Gen.\ PPL} \nonumber \\
&= \exp\!\Bigg(\!-\tfrac{1}{NL}\sum_{i=1}^N \sum_{t=1}^{L}\log p_{\text{GPT-2 Large}}\!\left(x^{(i)}_t \mid x^{(i)}_{<t}\right)\!\Bigg).
\end{align}
\par
}
\paragraph{Unigram entropy.}
\label{app:unigram-entropy}
Since a low Gen.\ PPL can be achieved by degenerate repetitive text, we also report the average unigram entropy of generated samples \citep{dieleman2022continuousdiffusioncategoricaldata}. Let $\mathcal{V}$ be the GPT-2 vocabulary and $c(v,\x^{(i)})$ the number of occurrences of token $v \in \mathcal{V}$ in sequence $\x^{(i)}$. We define the empirical unigram distribution $q^{(i)}(v)=c(v,\x^{(i)})/L$ and report
{\small
\begin{equation}
    \text{Entropy} \;=\; -\frac{1}{N}\sum_{i=1}^N \sum_{v\in\mathcal{V}} q^{(i)}(v)\,\log q^{(i)}(v).
\end{equation}
\par
}

\section{Algorithms}
\label{app:algorithms}
\paragraph{Stratified block size selection}
\algo{alg:stratified-block-size} samples one block size per GPU under the BlockGen mixture weights $\boldsymbol{\gamma}$, as described in \sec{paragraph:stratified-block-size-selection}.
\begin{algorithm}[t]
  \caption{Stratified Block Size Selection}
  \label{alg:stratified-block-size}
  \begin{algorithmic}[1]
  \REQUIRE Weights $\boldsymbol{\gamma} \in \Delta^M$, number of GPUs $D$
  \STATE $c_0 \gets 0$, \quad $c_n \gets \sum_{j=1}^{n} \gamma_j$ for $n = 1, \ldots, M$
  \STATE $u \sim \mathrm{Uniform}(0, 1)$
  \FOR{$d = 1$ \TO $D$}
      \STATE $\tilde{u}_d \gets \bigl(\tfrac{d-1}{D} + u\bigr) \mod 1$
      \STATE $i_d \gets n$ such that $c_{n-1} \leq \tilde{u}_d < c_n$
  \ENDFOR
  \STATE \textbf{return} $(i_1, \ldots, i_D)$
  \end{algorithmic}
\end{algorithm}
\paragraph{Block-level predictor-correctors}
\cref{alg:EIPC,alg:ARPC} list the EIPC and ARPC samplers for a single block. Lines that differ between the two samplers are highlighted in {\color{ourbrown}dark orange}. See \sec{sec:hybrid-sampling-with-blockgen} for the derivation.
\begin{figure*}[t]
\begin{minipage}[t]{0.48\linewidth}
\begin{algorithm}[H]
  \caption{EIPC sampler (one block)}
  \label{alg:EIPC}
  \begin{algorithmic}[1]
  \REQUIRE Prefix $\x^{<b}$, block size $L'$, denoiser $\hat{\x}_\theta$
  \STATE $\mathbf{z}^b \sim \mathrm{Prior}$
  \FOR{$i = 0$ \TO\ $N_\text{step}{-}1$}
    \STATE $t \gets (N_\text{step}{-}i)/N_\text{step}$, $s \gets (N_\text{step}{-}i{-}1)/N_\text{step}$
    \STATE $p^{\mathrm{D}} \gets \hat{\x}_\theta(\mathbf{z}^b, \x^{<b}, t; L')$
    \IF{\textsc{Informed}$(i)$}
      \STATE $\tilde{\x}^b \sim q_{0|t}(\cdot \mid \mathbf{z}^b, p^{\mathrm{D}})$
      \STATE \phantom{$p^{\mathrm{AR}} \gets \hat{\x}_\theta(\tilde{\x}^b, \x^{<b}; L'{=}1)$}
      \STATE \textcolor{ourbrown}{$\mathbf{g}^\ell \gets \sum_v p^{\mathrm{D},\ell}(v) \log p^{\mathrm{D},\ell}(v)$}
      \STATE $k \gets \mathrm{round}((1 - \alpha_s) L')$
      \STATE $\mathbf{m} \gets \textsc{BottomK}(\mathbf{g}, k)$
      \STATE $\mathbf{z}^b \gets \tilde{\x}^b$
      \STATE $\mathbf{z}^{b,\ell} \sim q_s(\cdot \mid \tilde{\x}^{b,\ell}; \alpha_s)$ for $\ell \in \mathbf{m}$
    \ELSE
      \STATE $\mathbf{z}^b \sim q_{s|t}(\cdot \mid \mathbf{z}^b, p^{\mathrm{D}})$
    \ENDIF
  \ENDFOR
  \STATE \textbf{return} $\mathbf{z}^b$
  \end{algorithmic}
\end{algorithm}
\end{minipage}\hfill
\begin{minipage}[t]{0.48\linewidth}
\begin{algorithm}[H]
  \caption{ARPC sampler (one block)}
  \label{alg:ARPC}
  \begin{algorithmic}[1]
  \REQUIRE Prefix $\x^{<b}$, block size $L'$, denoiser $\hat{\x}_\theta$
  \STATE $\mathbf{z}^b \sim \mathrm{Prior}$
  \FOR{$i = 0$ \TO\ $N_\text{step}{-}1$}
    \STATE $t \gets (N_\text{step}{-}i)/N_\text{step}$, $s \gets (N_\text{step}{-}i{-}1)/N_\text{step}$
    \STATE $p^{\mathrm{D}} \gets \hat{\x}_\theta(\mathbf{z}^b, \x^{<b}, t; L')$
    \IF{\textsc{Informed}$(i)$}
      \STATE $\tilde{\x}^b \sim q_{0|t}(\cdot \mid \mathbf{z}^b, p^{\mathrm{D}})$
      \STATE \textcolor{ourbrown}{$p^{\mathrm{AR}} \gets \hat{\x}_\theta(\tilde{\x}^b, \x^{<b}; L'{=}1)$}
      \STATE \textcolor{ourbrown}{$\mathbf{g}^\ell \gets \log p^{\mathrm{AR},\ell}(\tilde{\x}^{b,\ell})$}
      \STATE $k \gets \mathrm{round}((1 - \alpha_s) L')$
      \STATE $\mathbf{m} \gets \textsc{BottomK}(\mathbf{g}, k)$
      \STATE $\mathbf{z}^b \gets \tilde{\x}^b$
      \STATE $\mathbf{z}^{b,\ell} \sim q_s(\cdot \mid \tilde{\x}^{b,\ell}; \alpha_s)$ for $\ell \in \mathbf{m}$
    \ELSE
      \STATE $\mathbf{z}^b \sim q_{s|t}(\cdot \mid \mathbf{z}^b, p^{\mathrm{D}})$
    \ENDIF
  \ENDFOR
  \STATE \textbf{return} $\mathbf{z}^b$
  \end{algorithmic}
\end{algorithm}
\end{minipage}
\end{figure*}

\section{Additional Experimental Results}
\label{supp:additional-experiments}

\subsection{Raw GSM8K accuracies}
\label{app:raw-accuracies}
Tables at the end of this manuscript contain the raw GSM8K accuracies (models trained on TinyGSM, evaluated on the GSM8K test set) per NFE and temperature. First, we show the tables with block size 16 and then with block size 32. Within each group, we first show the results with ancestral sampling, then ARPC and finally EIPC. For ancestral and EIPC, we use single-block-size models, hence only ARPC uses the multi-block-size checkpoints. ARPC uses the mixture from \Eqn{eq:blockgen-mixture} with $\gamma_1 = 0.05$ and $\gamma_{L'} = 0.95$ for $L' \in \{16, 32\}$. Block size 16: \tab{tab:ancestral_variants_b16}, \tab{tab:arpc_variants_b16}, \tab{tab:eipc_variants_b16}. Block size 32: \tab{tab:ancestral_variants_b32}, \tab{tab:arpc_variants_b32}, \tab{tab:eipc_variants_b32}.

\paragraph{Counting NFE for ARPC}
Each ARPC informed step performs two denoiser forward passes. First, we sample from $q_{0|t}$ and then compute the NLL in AR mode ($L'{=}1$, \algo{alg:ARPC}) to decide which positions to re-mask / re-inject noise. EIPC and ancestral steps only need a single forward pass. When comparing samplers, we account for both the AR and diffusion mode predictions for ARPC, such that the sampling time is roughly equivalent between ancestral, EIPC, and ARPC when matching NFE. The first $n_\text{warmup}$ steps are ancestral-only. Within the remaining $N_\text{step}-n_\text{warmup}$ iterations, an informed step occurs every $\text{GE}$ steps, starting at the first non-warmup step. The number of informed steps is therefore $\lfloor(N_\text{step}-1-n_\text{warmup})/\text{GE}\rfloor+1$, and adding the $N_\text{step}$ predictor passes gives
{\small
\begin{equation}
  \label{eq:arpc-nfe}
  \mathrm{NFE} = N_\text{step} + \lfloor(N_\text{step}-1-n_\text{warmup})/\text{GE}\rfloor + 1.
\end{equation}
\par
}
For example, from \tab{tab:arpc_variants_b16}: $(N_\text{step},\text{GE},n_\text{warmup})=(3,3,0)$ has one informed step at $i{=}0$, so $\mathrm{NFE}=3+0+1=4$. $(12,3,0)$ has four informed steps at $i\in\{0,3,6,9\}$, so $\mathrm{NFE}=12+3+1=16$. $(14,4,8)$ has eight ancestral warmup steps then two informed steps at $i\in\{8,12\}$, so $\mathrm{NFE}=14+1+1=16$.

\paragraph{Block size 16 vs.\ 32 at matched total NFE}
\tab{tab:single_block_b16_vs_b32} compares single-block models at block sizes 16 and 32 under a matched total-NFE budget of 1024 per 512-token sequence (32 NFE/block for b16 across 32 blocks, 64 NFE/block for b32 across 16 blocks). At equal compute, the block-16 model outperforms the block-32 model on every reported (noising prior, sampler) combination.

\subsection{Validation Perplexity on OpenWebText}
\label{app:owt-val-ppl}
\tab{tab:valid-ppl-after-250k-steps} reports the validation perplexity for single-block-size models after 250k training steps on LM1B and OWT.

\paragraph{Training objectives for single-block BDMs}
For Masked BDMs, BD3-LM \citep{arriola2025blockdiffusioninterpolatingautoregressive} uses a costly variance reduction scheme that optimizes the noise schedule every 5k steps over the validation set. We find that training with unweighted cross-entropy matches their performance on OWT and slightly improves it on LM1B, without the overhead. We therefore use the unweighted CE for all masked models, which recent work showed optimizes the true NELBO \citep{shi2025demystifyingdiffusionobjectivesreweighted,sahoo2026scalingbeyondmasked,sahoo2025esotericlanguagemodels}. For uniform-state BDMs, the unweighted CE underperforms the NELBO on OWT for block sizes $>4$, so we use the NELBO for all $L'>1$ \citep{vonrütte2025generalizedinterpolatingdiscretediffusion}. Consistent with prior work \citep{schiff2025simpleguidancemechanismsdiscrete, sahoo2025diffusionduality}, uniform-state BDMs slightly underperform masked ones in likelihood, yet still outperform the full-sequence USDM Duo \citep{sahoo2025diffusionduality}.

\subsection{Additional Figures on OpenWebText}
\label{app:owt-additional-figures}
The figures in this subsection appear at the end of the manuscript, after the TinyGSM figures and before the tables. All figures plot Generative Perplexity (GPT-2 Large evaluator) against per-sample unigram entropy, with each curve a temperature sweep at fixed per-block NFE. Lower Gen.\ PPL at matched entropy is better.

\paragraph{Sampling protocol}
We sample $1024$ sequences of length $1024$ per combination of (sampler, prior, $T$, NFE) across the four samplers. Single-block ancestral and EIPC use the single-block-$16$ diffusion model. ARPC uses the BlockGen checkpoint with $\gamma_1{=}0.05$, $\gamma_{16}{=}0.95$: this checkpoint has higher perplexity than the uniform-$\bm{\gamma}$ one, but ARPC samples tokens from the $L'{=}16$ predictions and uses the $L'{=}1$ (AR) pass only as a verifier, so the larger mass on $\gamma_{16}$ is more suited for ARPC.

\paragraph{EIPC as a confidence-based corrector}
EIPC re-noises tokens whose marginals carry the highest entropy and uses no AR auxiliary, so it applies to both priors. On masked diffusion, EIPC matches the single-block ancestral sampler while ARPC reaches a better Gen.\ PPL frontier. On uniform diffusion, EIPC performs better than ARPC, consistent with the GSM8K observation at low temperature.

\begin{figure*}[t]
  \centering
  \begin{subfigure}[t]{0.49\textwidth}
    \centering
    \includegraphics[width=\textwidth]{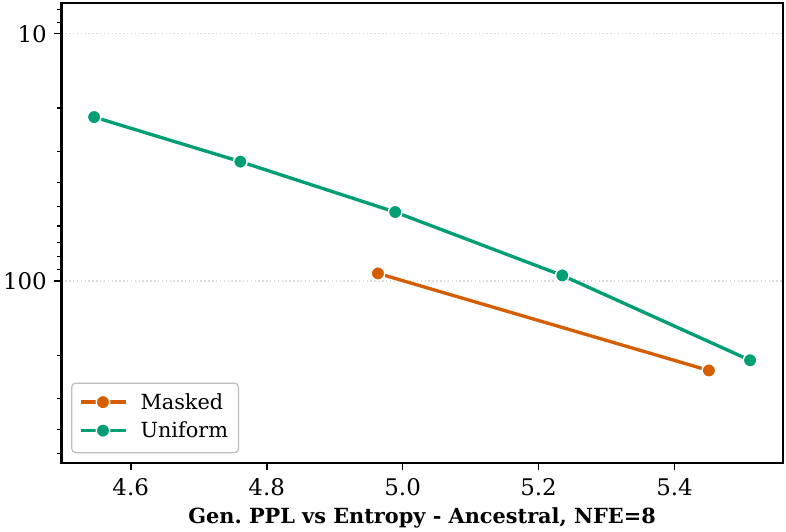}
    \label{fig:owt-ancestral-mvu-nfe8}
  \end{subfigure}\hfill
  \begin{subfigure}[t]{0.49\textwidth}
    \centering
    \includegraphics[width=\textwidth]{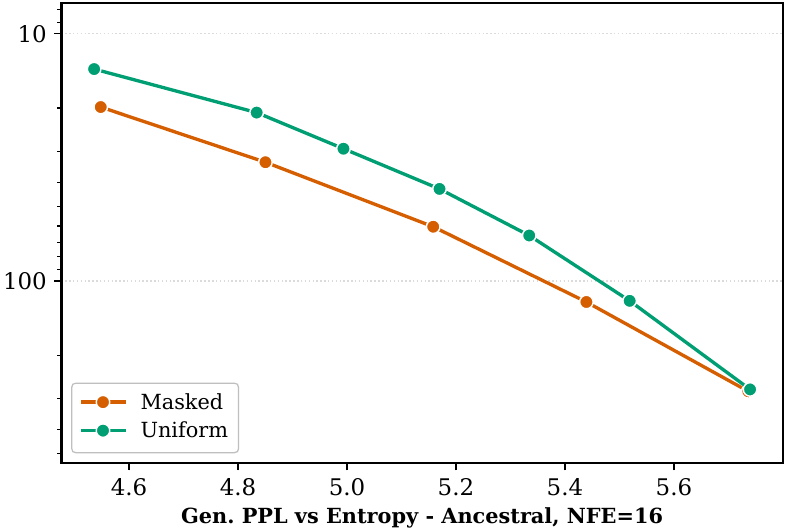}
    \label{fig:owt-ancestral-mvu-nfe16}
  \end{subfigure}

  \vspace{0.5em}
  \begin{subfigure}[t]{0.49\textwidth}
    \centering
    \includegraphics[width=\textwidth]{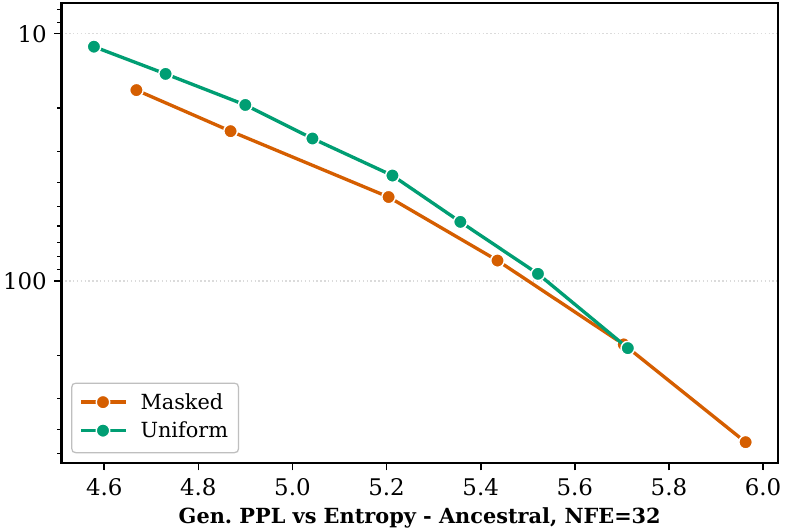}
    \label{fig:owt-ancestral-mvu-nfe32}
  \end{subfigure}\hfill
  \begin{subfigure}[t]{0.49\textwidth}
    \centering
    \includegraphics[width=\textwidth]{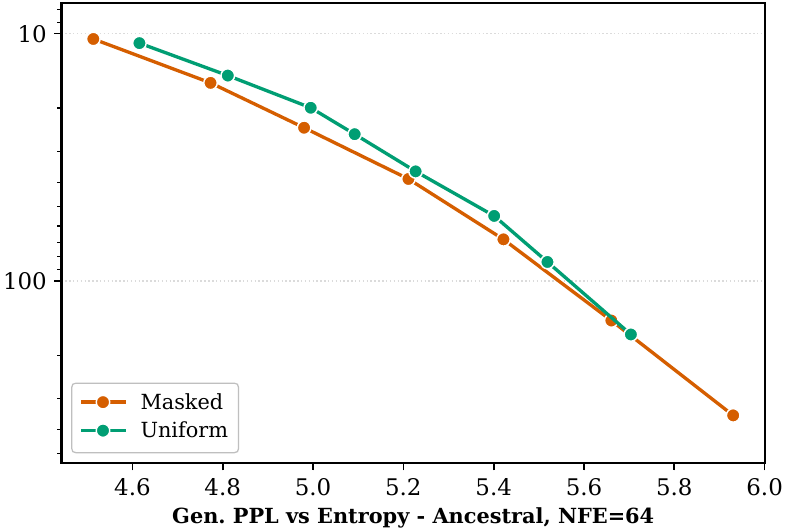}
    \label{fig:owt-ancestral-mvu-nfe64}
  \end{subfigure}
  \caption{\textbf{Single-block ancestral sampling on OpenWebText: masked vs uniform diffusion} at per-block NFE $\in\{8, 16, 32, 64\}$. Uniform reaches a lower Gen.\ PPL frontier than masked at every NFE, with the gap narrowing as NFE grows.}
  \label{fig:owt-ancestral-mvu}
\end{figure*}

\paragraph{Ancestral sampling: uniform vs masked}
\fig{fig:owt-ancestral-mvu} compares masked and uniform diffusion under single-block ancestral sampling at four per-block NFE budgets ($8, 16, 32, 64$). Uniform reaches a lower Gen.\ PPL frontier than masked across all budgets, with the gap narrowing at higher NFE.

\begin{figure*}[t]
  \centering
  \begin{subfigure}[t]{0.49\textwidth}
    \centering
    \includegraphics[width=\textwidth]{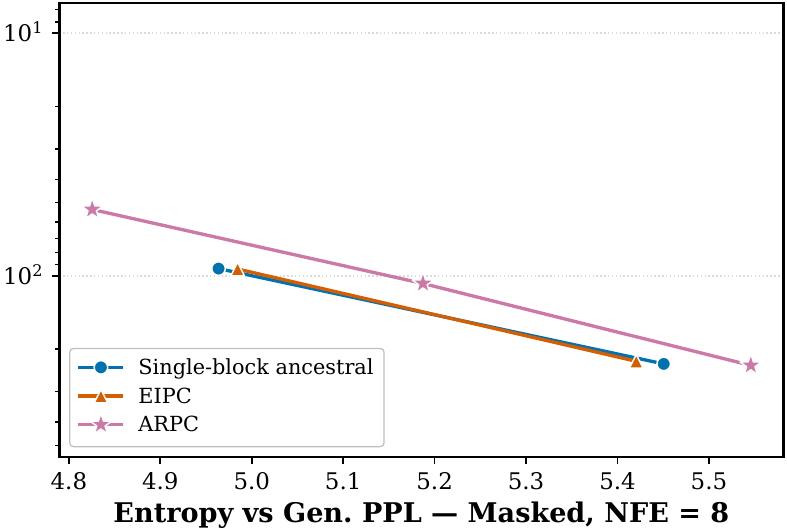}
    \label{fig:owt-blockmethods-masked-nfe8}
  \end{subfigure}\hfill
  \begin{subfigure}[t]{0.49\textwidth}
    \centering
    \includegraphics[width=\textwidth]{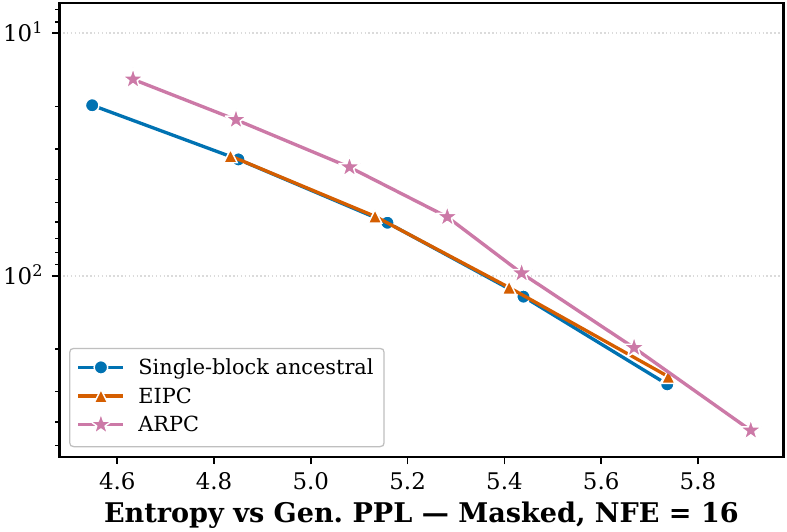}
    \label{fig:owt-blockmethods-masked-nfe16}
  \end{subfigure}

  \vspace{0.5em}
  \begin{subfigure}[t]{0.49\textwidth}
    \centering
    \includegraphics[width=\textwidth]{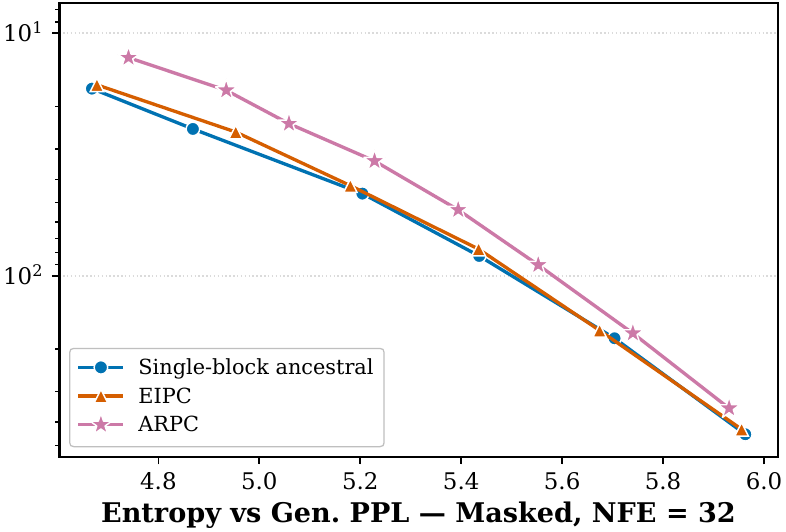}
    \label{fig:owt-blockmethods-masked-nfe32}
  \end{subfigure}\hfill
  \begin{subfigure}[t]{0.49\textwidth}
    \centering
    \includegraphics[width=\textwidth]{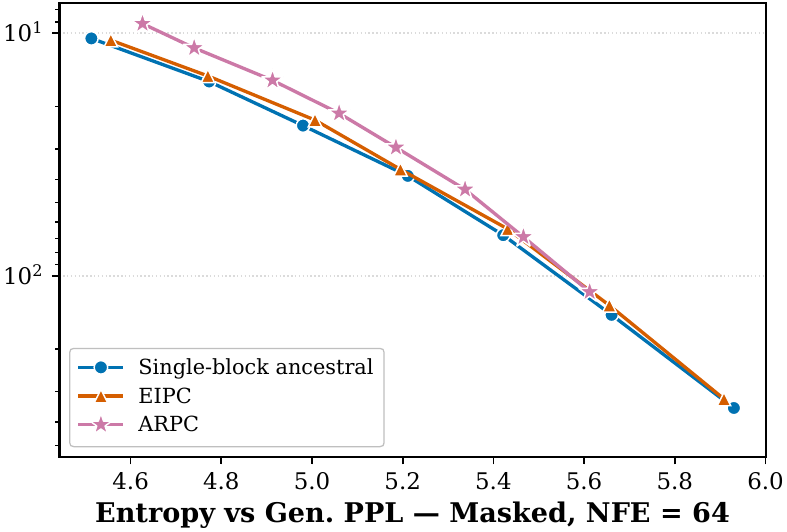}
    \label{fig:owt-blockmethods-masked-nfe64}
  \end{subfigure}
  \caption{\textbf{Block-level samplers on OpenWebText, masked prior:} single-block ancestral, EIPC, and ARPC at per-block NFE $\in\{8, 16, 32, 64\}$. ARPC reaches the lowest Gen.\ PPL frontier at every NFE.}
  \label{fig:owt-blockmethods-masked}
\end{figure*}

\paragraph{Block-level methods (masked prior)}
\fig{fig:owt-blockmethods-masked} compares single-block ancestral, EIPC, and ARPC under the masked prior at four per-block NFE budgets. ARPC reaches the lowest Gen.\ PPL frontier across all NFE values.

\begin{figure*}[t]
  \centering
  \begin{subfigure}[t]{0.49\textwidth}
    \centering
    \includegraphics[width=\textwidth]{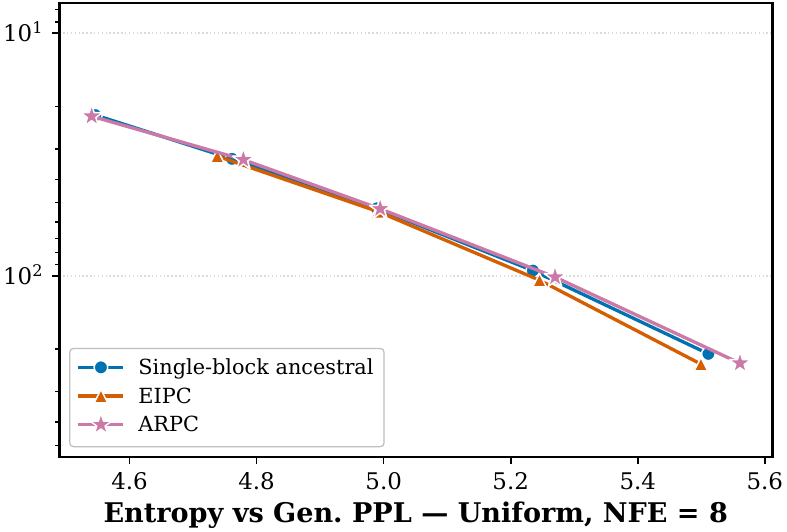}
    \label{fig:owt-blockmethods-uniform-nfe8}
  \end{subfigure}\hfill
  \begin{subfigure}[t]{0.49\textwidth}
    \centering
    \includegraphics[width=\textwidth]{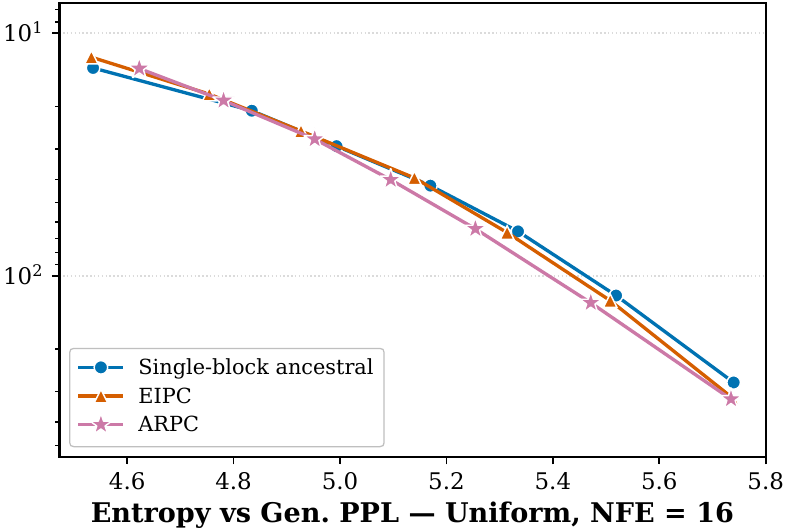}
    \label{fig:owt-blockmethods-uniform-nfe16}
  \end{subfigure}

  \vspace{0.5em}
  \begin{subfigure}[t]{0.49\textwidth}
    \centering
    \includegraphics[width=\textwidth]{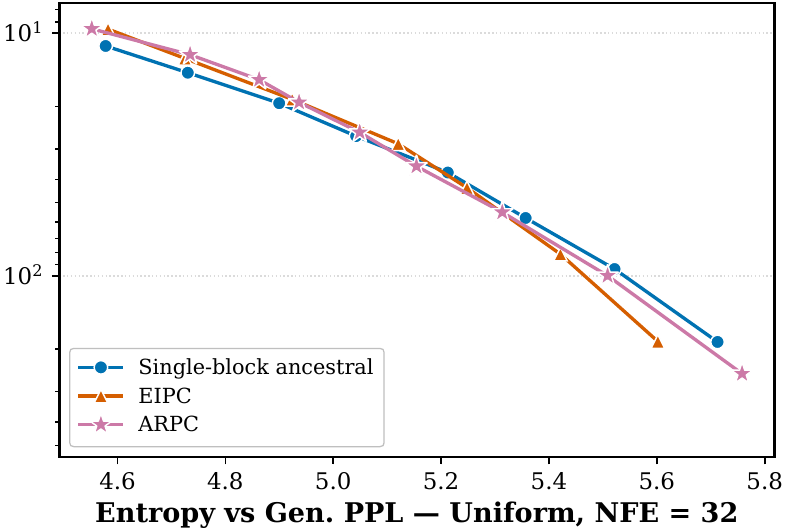}
    \label{fig:owt-blockmethods-uniform-nfe32}
  \end{subfigure}\hfill
  \begin{subfigure}[t]{0.49\textwidth}
    \centering
    \includegraphics[width=\textwidth]{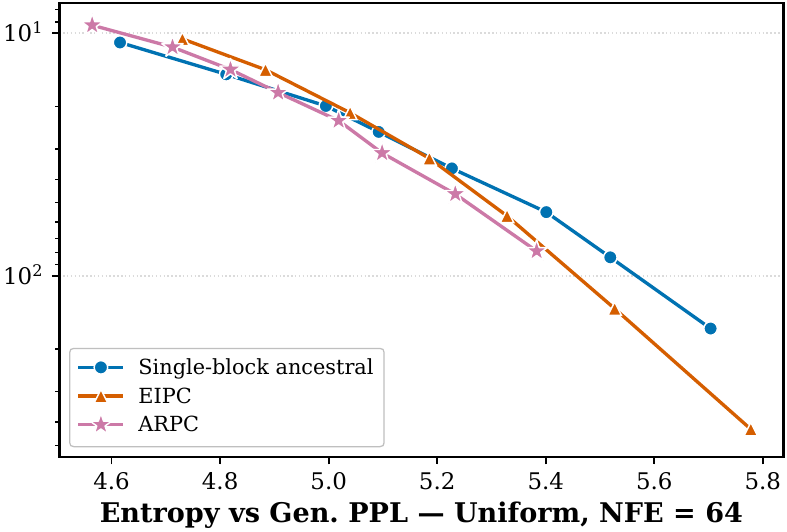}
    \label{fig:owt-blockmethods-uniform-nfe64}
  \end{subfigure}
  \caption{\textbf{Block-level samplers on OpenWebText, uniform prior:} single-block ancestral, EIPC, and ARPC at per-block NFE $\in\{8, 16, 32, 64\}$. ARPC and EIPC trade places across temperature and NFE, and both remain close to single-block ancestral.}
  \label{fig:owt-blockmethods-uniform}
\end{figure*}

\paragraph{Block-level methods (uniform prior)}
\fig{fig:owt-blockmethods-uniform} reports the same comparison under the uniform prior. ARPC and EIPC trade places across temperature and NFE, and both stay close to single-block ancestral. The relative ordering depends on the regime.

\begin{figure*}[t]
  \centering
  \begin{subfigure}[t]{0.49\textwidth}
    \centering
    \includegraphics[width=\textwidth]{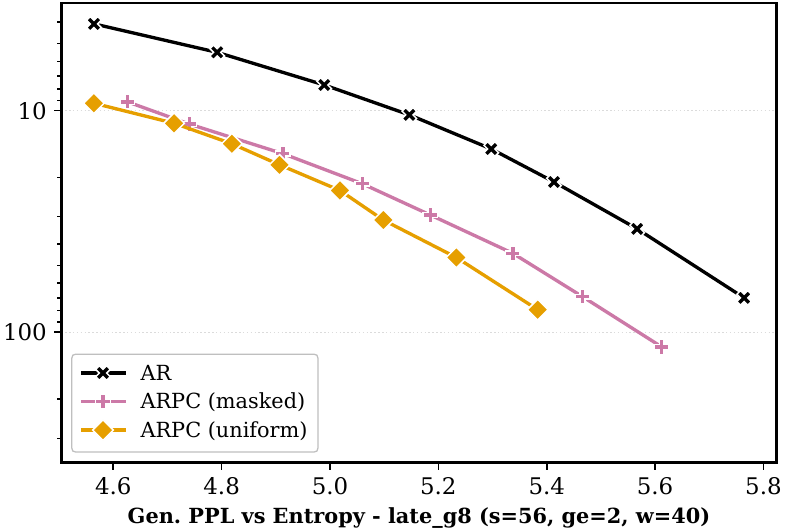}
    \label{fig:owt-arpc-late-g8}
  \end{subfigure}\hfill
  \begin{subfigure}[t]{0.49\textwidth}
    \centering
    \includegraphics[width=\textwidth]{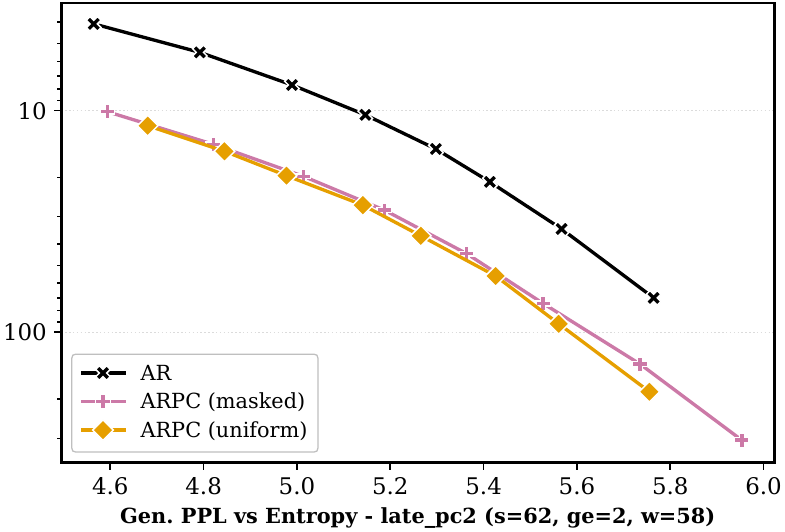}
    \label{fig:owt-arpc-late-pc2}
  \end{subfigure}

  \vspace{0.5em}
  \begin{subfigure}[t]{0.49\textwidth}
    \centering
    \includegraphics[width=\textwidth]{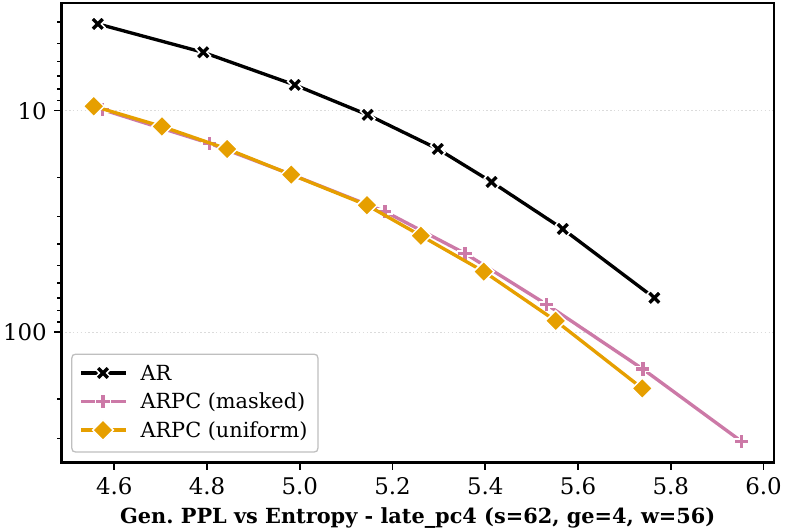}
    \label{fig:owt-arpc-late-pc4}
  \end{subfigure}
  \caption{\textbf{ARPC vs AR on OpenWebText at per-block NFE $=64$, additional late-correction schedules.} From left to right: a wider corrector spacing ($\text{GE}=8$, with a $40$-step warmup), and two schedules with very few correctors at the end ($2$ correctors after a $58$-step warmup; $1$ corrector after a $56$-step warmup with $\text{GE}=4$). AR retains lower Gen.\ PPL across the practical temperature range in all three settings, and masked-ARPC sits closer to AR than uniform-ARPC.}
  \label{fig:owt-arpc-late-schedules}
\end{figure*}

\paragraph{ARPC at NFE per block $=64$ across late-correction schedules}
\fig{fig:owt-arpc-late-schedules} reports three additional late-correction schedules at per-block NFE $=64$, complementing the right panel of \fig{fig:owt-arpc-mvu} in the main text. Across all three schedules, AR retains lower Gen.\ PPL than ARPC at the practical temperatures, and masked-ARPC remains closer to AR than uniform-ARPC.

\begin{figure*}[t]
  \centering
  \includegraphics[width=0.6\textwidth]{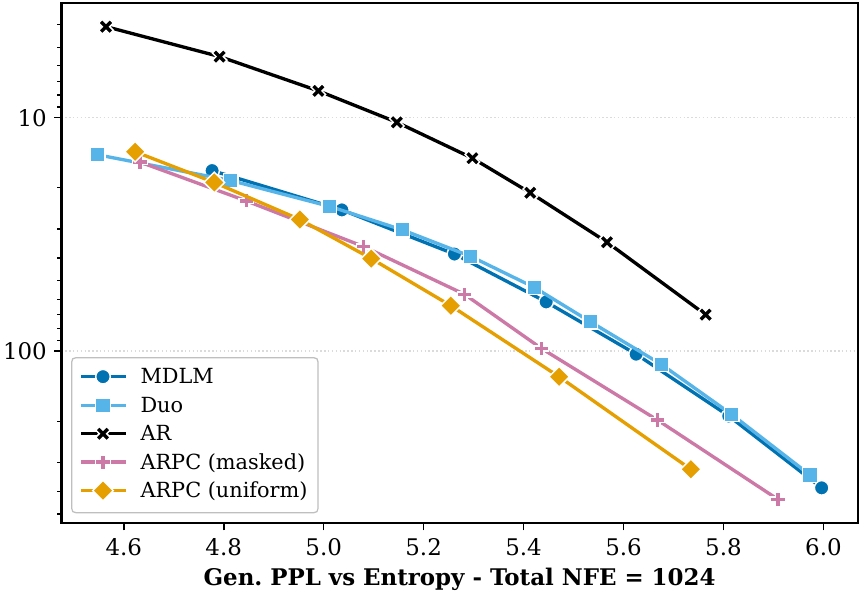}
  \caption{\textbf{Matched total-NFE frontier on OpenWebText, total NFE $=1024$.} AR runs at one forward pass per token; MDLM and Duo are full-sequence diffusion samplers at $1024$ ancestral steps; ARPC (masked) and ARPC (uniform) run block-by-block at $16$ NFE per block across $64$ blocks. At matched total compute, single-block samplers reach lower Gen.\ PPL than block-by-block ARPC, consistent with the trend reported for masked diffusion \citep{sahoo2025esotericlanguagemodels}.}
  \label{fig:owt-frontier-matched}
\end{figure*}

\paragraph{Matched total-NFE comparison}
\fig{fig:owt-frontier-matched} reports the Gen.\ PPL / entropy frontier at matched total NFE $=1024$. AR uses one forward pass per token. MDLM and Duo are full-sequence diffusion samplers run with $1024$ ancestral steps; ARPC (masked) and ARPC (uniform) are block-by-block at $32$ NFE per block over $32$ blocks. At matched total NFE, single-block samplers reach lower Gen.\ PPL than block-by-block ARPC, consistent with prior reports for masked diffusion \citep{sahoo2025esotericlanguagemodels} and extending the observation to uniform diffusion. Because BlockGen allows KV-caching, matching the NFEs does not mean matching the throughput, and BlockGen is generally faster than full-sequence diffusion during sampling.

\subsection{Additional GSM8K Figures}
\label{app:sampler-ablations}
The figures in this subsection appear at the end of the manuscript, before the tables.

\begin{figure*}[t]
  \centering
  \begin{subfigure}[t]{0.49\textwidth}
    \centering
    \includegraphics[width=\textwidth]{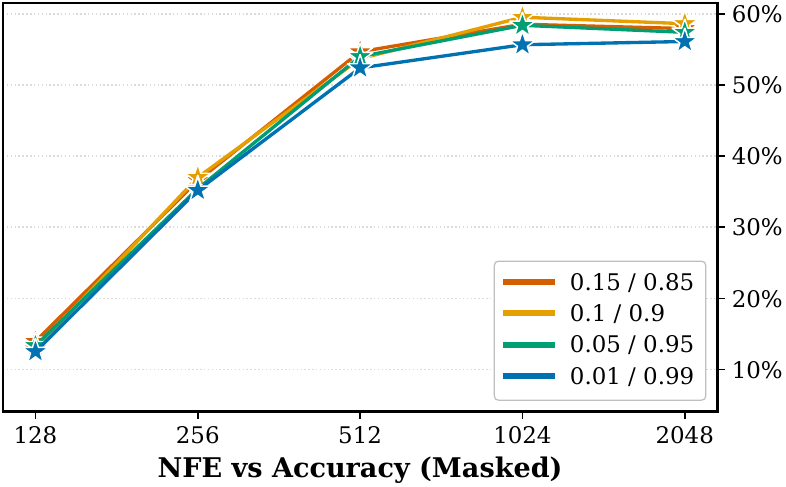}
    \label{fig:arpc_weights_b16_masked}
  \end{subfigure}\hfill
  \begin{subfigure}[t]{0.49\textwidth}
    \centering
    \includegraphics[width=\textwidth]{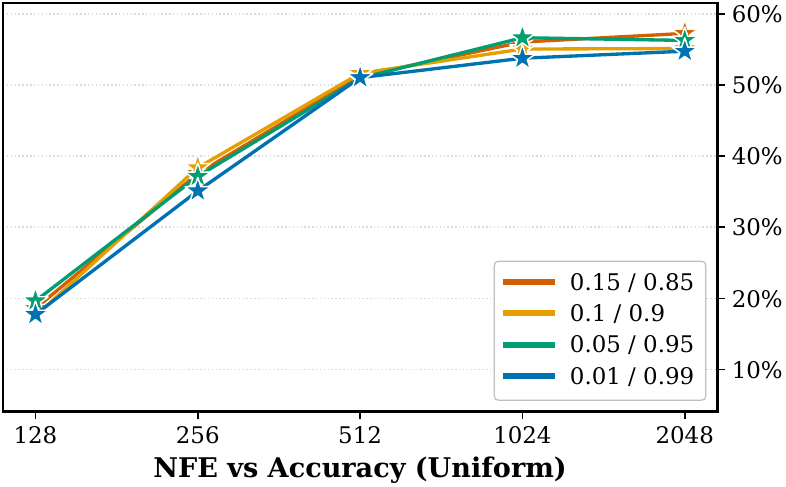}
    \label{fig:arpc_weights_b16_uniform}
  \end{subfigure}
  \caption{\textbf{Effect of mixture weights on ARPC with block size 16} as a function of NFE, at $T{=}1$. Each curve shows the best ARPC performance for a given NFE, one curve per value of $\gamma_1$ in \Eqn{eq:blockgen-mixture}, with $\gamma_{16} = 1 - \gamma_1$. Masked diffusion (left) and uniform diffusion (right).}
  \label{fig:arpc_weights_b16}
\end{figure*}

\paragraph{Effect of mixture weights at block size 16}
\fig{fig:arpc_weights_b16} shows the accuracy of ARPC at $T{=}1$ across four values of $\gamma_1$ in \Eqn{eq:blockgen-mixture}, with $\gamma_{16} = 1 - \gamma_1$.

\begin{figure*}[t]
  \centering
  \begin{subfigure}[t]{0.49\textwidth}
    \centering
    \includegraphics[width=\textwidth]{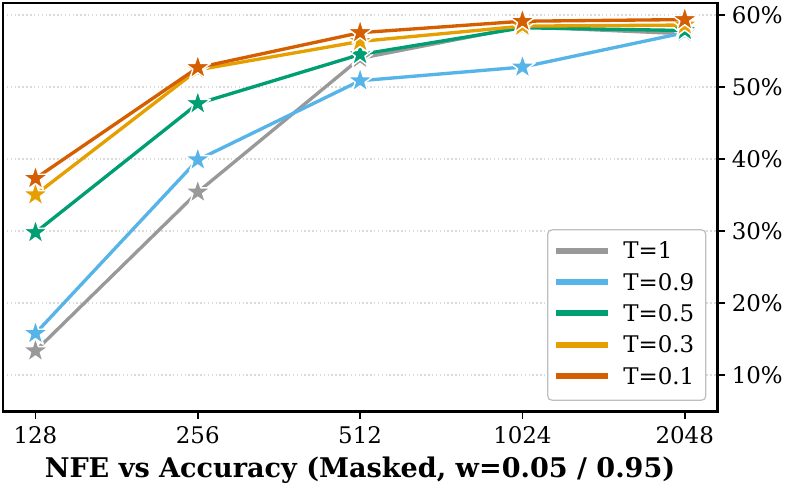}
    \label{fig:arpc_temps_b16_masked}
  \end{subfigure}\hfill
  \begin{subfigure}[t]{0.49\textwidth}
    \centering
    \includegraphics[width=\textwidth]{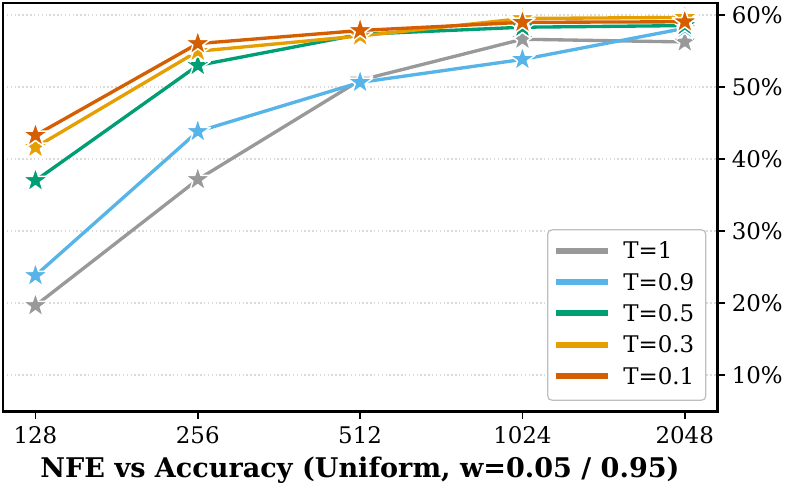}
    \label{fig:arpc_temps_b16_uniform}
  \end{subfigure}
  \caption{\textbf{Effect of sampling temperature on ARPC with block size 16} as a function of NFE. Each curve shows the best ARPC performance for a given NFE, one curve per sampling temperature ($T \in \{1.0, 0.9, 0.5, 0.3, 0.1\}$). ARPC uses the multi-block mixture from \Eqn{eq:blockgen-mixture} with $\gamma_1 = 0.05$, $\gamma_{16} = 0.95$. Masked diffusion (left) and uniform diffusion (right).}
  \label{fig:arpc_temps_b16}
\end{figure*}

\paragraph{Effect of sampling temperature at block size 16}
\fig{fig:arpc_temps_b16} shows the accuracy of ARPC at $T \in \{1.0, 0.9, 0.5, 0.3, 0.1\}$, using \Eqn{eq:blockgen-mixture} with $\gamma_1 = 0.05$ and $\gamma_{16} = 0.95$.

\begin{figure*}[t]
  \centering
  \begin{subfigure}[t]{0.49\textwidth}
    \centering
    \includegraphics[width=\textwidth]{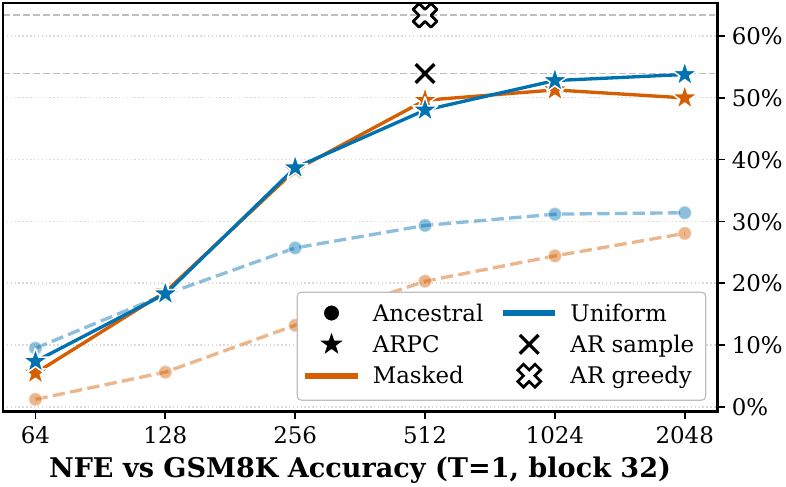}
    \label{fig:ar_vs_ancestral_vs_arpc_b32_T1}
  \end{subfigure}\hfill
  \begin{subfigure}[t]{0.49\textwidth}
    \centering
    \includegraphics[width=\textwidth]{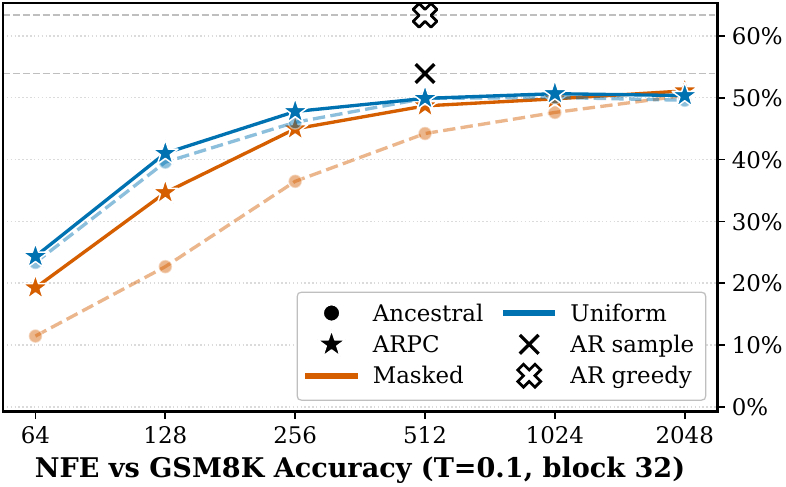}
    \label{fig:ar_vs_ancestral_vs_arpc_b32_T0p1}
  \end{subfigure}
  \caption{\textbf{GSM8K accuracy with block size 32} as a function of NFE. Models are trained on TinyGSM and evaluated on the GSM8K test set. Each curve shows the best performance for a given NFE. ARPC uses checkpoints trained with the mixture in \Eqn{eq:blockgen-mixture}, with $\gamma_1 = 0.05$, $\gamma_{32} = 0.95$, and ancestral uses a single-block-size model. Dashed lines show the AR baseline ($53.9\%$ sampled, $63.3\%$ greedy). ARPC outperforms ancestral across the NFE range (left and right).}
  \label{fig:ar_vs_ancestral_vs_arpc_b32}
\end{figure*}

\paragraph{AR vs ancestral vs ARPC at block size 32}
\fig{fig:ar_vs_ancestral_vs_arpc_b32} compares AR, ancestral and ARPC at block size 32. ARPC uses \Eqn{eq:blockgen-mixture} with $\gamma_1 = 0.05$ and $\gamma_{32} = 0.95$, and ancestral uses the single-block-32 model.

\begin{figure*}[t]
  \centering
  \begin{subfigure}[t]{0.49\textwidth}
    \centering
    \includegraphics[width=\textwidth]{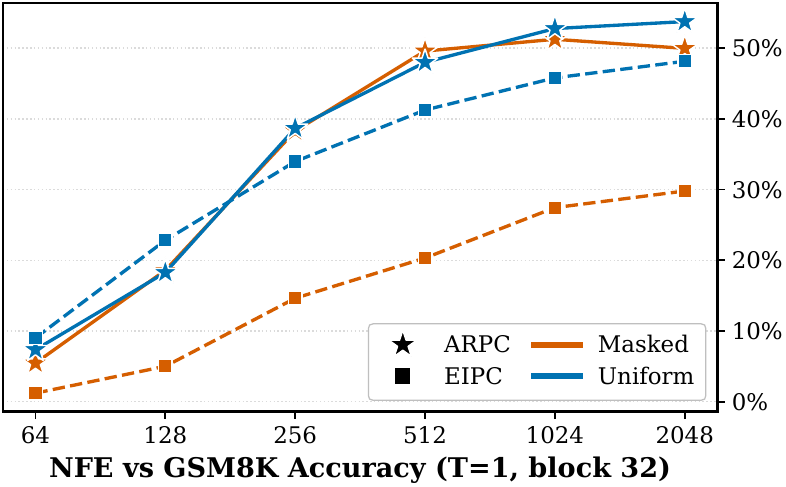}
    \label{fig:arpc_vs_eipc_b32_T1}
  \end{subfigure}\hfill
  \begin{subfigure}[t]{0.49\textwidth}
    \centering
    \includegraphics[width=\textwidth]{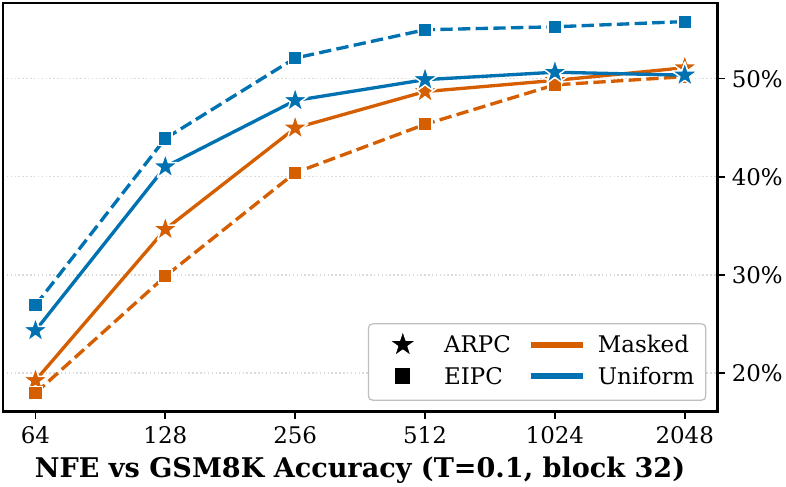}
    \label{fig:arpc_vs_eipc_b32_T0p1}
  \end{subfigure}
  \caption{\textbf{ARPC vs EIPC GSM8K accuracy with block size 32} as a function of NFE. Models are trained on TinyGSM and evaluated on the GSM8K test set. Each curve shows the best performance for a given NFE. EIPC uses a single-block-32 model, and ARPC uses the multi-block mixture from \Eqn{eq:blockgen-mixture} with $\gamma_1 = 0.05$, $\gamma_{32} = 0.95$. ARPC outperforms EIPC across most NFE budgets (left and right).}
  \label{fig:arpc_vs_eipc_b32}
\end{figure*}

\paragraph{ARPC vs EIPC at block size 32}
\fig{fig:arpc_vs_eipc_b32} compares ARPC and EIPC at block size 32. ARPC uses \Eqn{eq:blockgen-mixture} with $\gamma_1 = 0.05$ and $\gamma_{32} = 0.95$, and EIPC uses the single-block-32 model.

\begin{figure*}[t]
  \centering
  \begin{subfigure}[t]{0.49\textwidth}
    \centering
    \includegraphics[width=\textwidth]{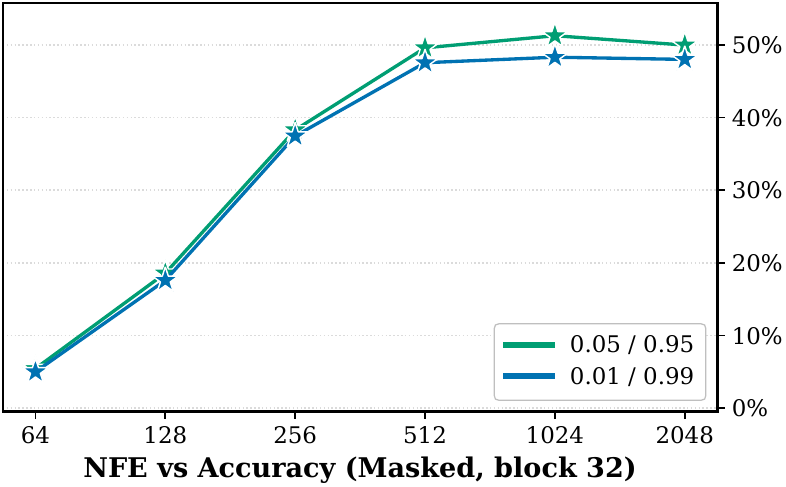}
    \label{fig:arpc_weights_b32_masked}
  \end{subfigure}\hfill
  \begin{subfigure}[t]{0.49\textwidth}
    \centering
    \includegraphics[width=\textwidth]{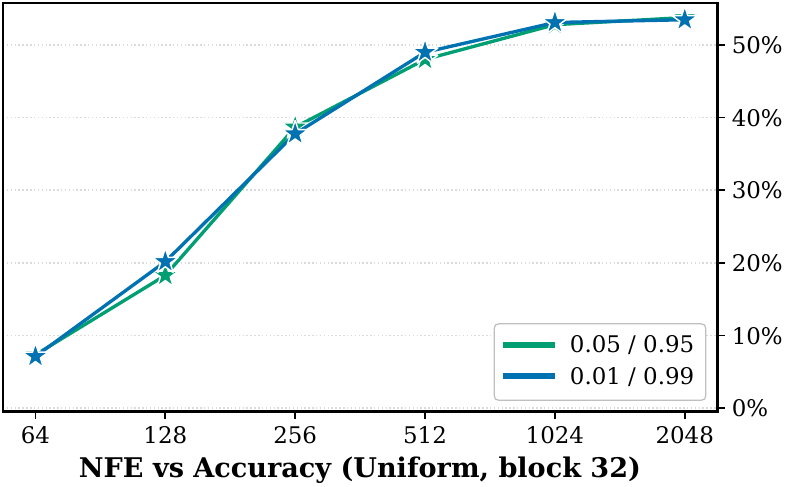}
    \label{fig:arpc_weights_b32_uniform}
  \end{subfigure}
  \caption{\textbf{Effect of mixture weights on ARPC with block size 32} as a function of NFE, at $T{=}1$. Each curve shows the best ARPC performance for a given NFE, one curve per $(\gamma_1, \gamma_{32}) \in \{(0.05, 0.95), (0.01, 0.99)\}$ in \Eqn{eq:blockgen-mixture}. Masked diffusion (left) and uniform diffusion (right).}
  \label{fig:arpc_weights_b32}
\end{figure*}

\paragraph{Effect of mixture weights at block size 32}
\fig{fig:arpc_weights_b32} shows the accuracy of ARPC at $T{=}1$ for $(\gamma_1, \gamma_{32}) \in \{(0.05, 0.95), (0.01, 0.99)\}$ in \Eqn{eq:blockgen-mixture}.

\begin{figure*}[t]
  \centering
  \begin{subfigure}[t]{0.49\textwidth}
    \centering
    \includegraphics[width=\textwidth]{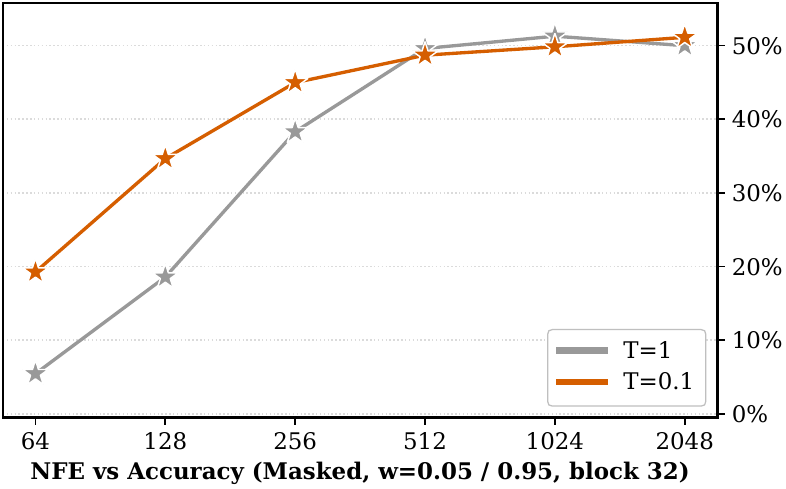}
    \label{fig:arpc_temps_b32_masked}
  \end{subfigure}\hfill
  \begin{subfigure}[t]{0.49\textwidth}
    \centering
    \includegraphics[width=\textwidth]{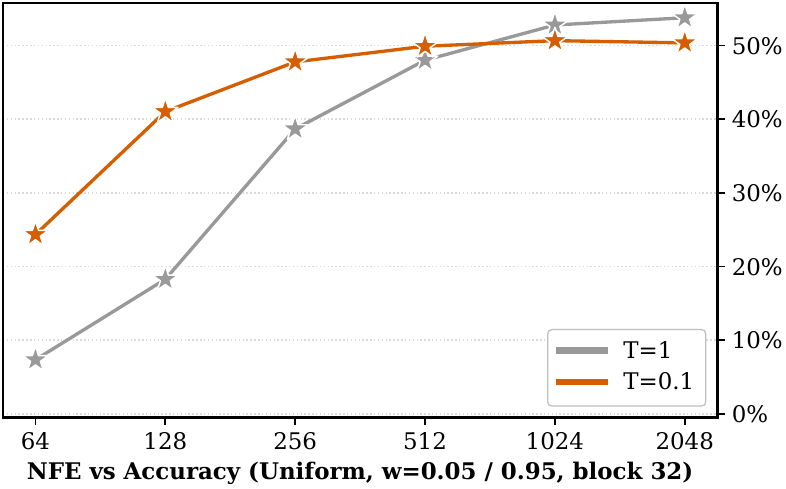}
    \label{fig:arpc_temps_b32_uniform}
  \end{subfigure}
  \caption{\textbf{Effect of sampling temperature on ARPC with block size 32} as a function of NFE. Each curve shows the best ARPC performance for a given NFE, one curve per sampling temperature ($T \in \{1, 0.1\}$). ARPC uses the multi-block mixture from \Eqn{eq:blockgen-mixture} with $\gamma_1 = 0.05$, $\gamma_{32} = 0.95$. Masked diffusion (left) and uniform diffusion (right).}
  \label{fig:arpc_temps_b32}
\end{figure*}

\paragraph{Effect of sampling temperature at block size 32}
\fig{fig:arpc_temps_b32} shows the accuracy of ARPC at $T \in \{1, 0.1\}$, using \Eqn{eq:blockgen-mixture} with $\gamma_1 = 0.05$ and $\gamma_{32} = 0.95$.

\begin{figure*}[t]
  \centering
  \begin{subfigure}[t]{0.49\textwidth}
    \centering
    \includegraphics[width=\textwidth]{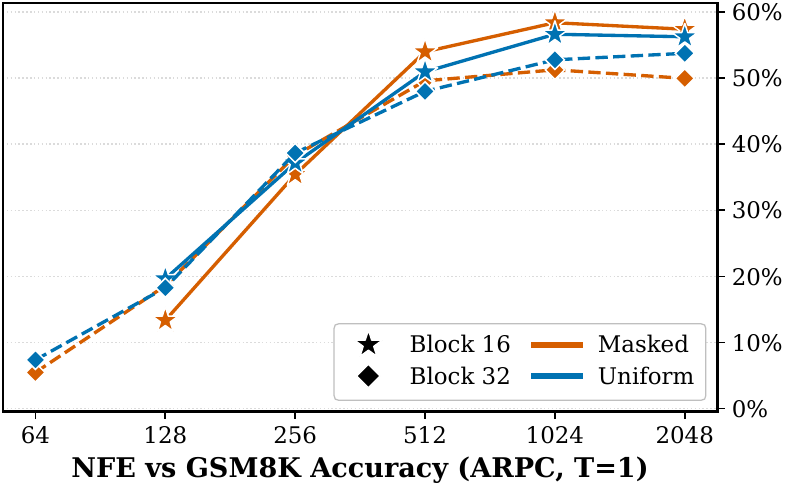}
    \label{fig:arpc_b16_vs_b32_T1}
  \end{subfigure}\hfill
  \begin{subfigure}[t]{0.49\textwidth}
    \centering
    \includegraphics[width=\textwidth]{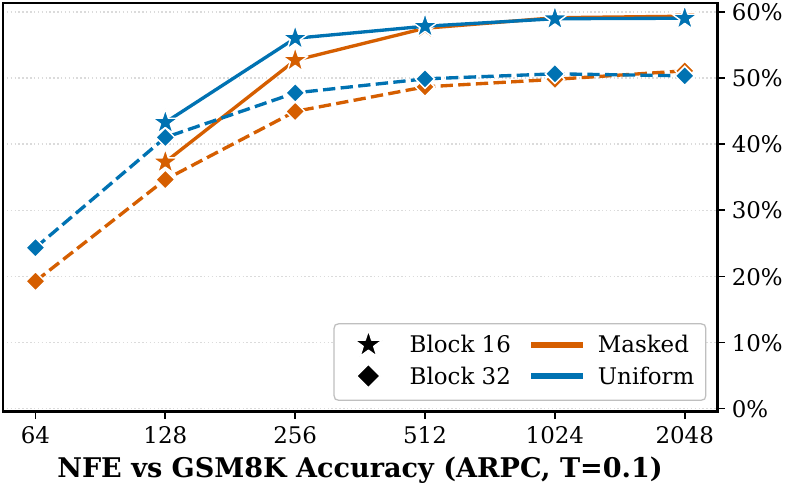}
    \label{fig:arpc_b16_vs_b32_T0p1}
  \end{subfigure}
  \caption{\textbf{ARPC GSM8K accuracy at block sizes 16 vs 32} as a function of NFE per block. Models are trained on TinyGSM and evaluated on the GSM8K test set. Each curve shows the best ARPC performance for a given NFE. Both curves use the mixture from \Eqn{eq:blockgen-mixture} with $\gamma_1 = 0.05$ and $\gamma_{L} = 0.95$ for $L \in \{16, 32\}$. Block size 16 reaches higher accuracy than block size 32 at matched NFE per block (left and right).}
  \label{fig:arpc_b16_vs_b32}
\end{figure*}

\begin{figure*}[t]
  \centering
  \begin{subfigure}[t]{0.49\textwidth}
    \centering
    \includegraphics[width=\textwidth]{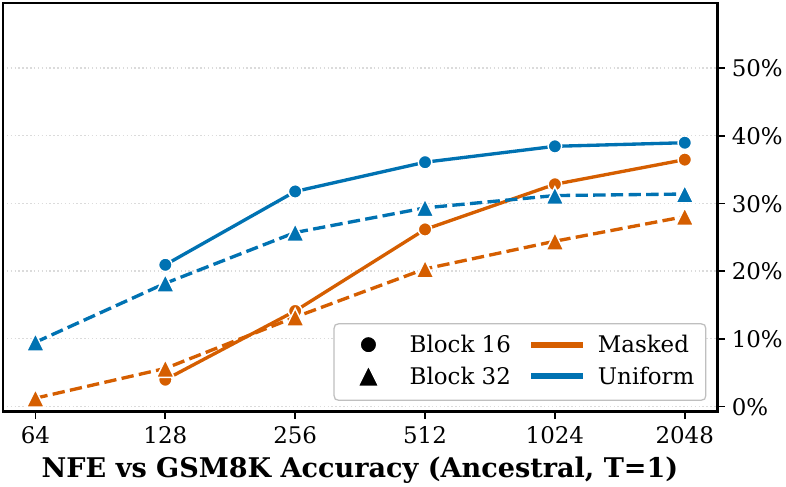}
    \label{fig:ancestral_b16_vs_b32_T1}
  \end{subfigure}\hfill
  \begin{subfigure}[t]{0.49\textwidth}
    \centering
    \includegraphics[width=\textwidth]{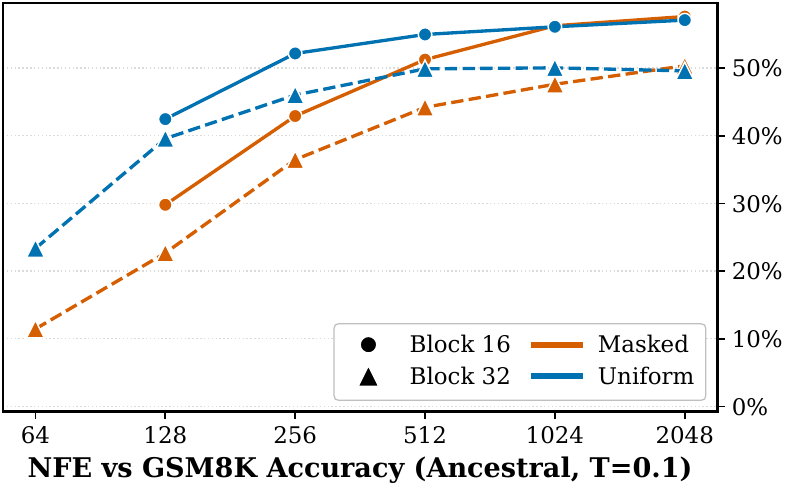}
    \label{fig:ancestral_b16_vs_b32_T0p1}
  \end{subfigure}
  \caption{\textbf{Ancestral sampling GSM8K accuracy at block sizes 16 vs 32} as a function of NFE per block. Models are trained on TinyGSM and evaluated on the GSM8K test set. Each curve shows the accuracy for a given NFE on the single-block-16 and single-block-32 models. Block size 16 reaches higher accuracy than block size 32 at matched NFE per block (left and right).}
  \label{fig:ancestral_b16_vs_b32}
\end{figure*}

\paragraph{Block size 16 vs 32}
\fig{fig:arpc_b16_vs_b32} compares ARPC at block sizes 16 and 32 on a shared NFE-per-block axis. \fig{fig:ancestral_b16_vs_b32} compares ancestral sampling at block sizes 16 and 32 on the same axis.

\section{Computing the Validation Perplexity}
\label{supp:valid-ppl}
This section defines how we compute the validation perplexity.
\paragraph{Token- and sequence-level evaluation}
Let the validation set contain $J$ sequences, indexed by $j\in\{1,\dots,J\}$. Each sequence is represented as $\x_{(j)}\in\mathcal{V}^{L}$ with an associated binary mask $\mathbf{a}_{(j)}\in\{0,1\}^{L}$, where $a_{(j),\ell}=1$ indicates that position $\ell$ is not padding.
We evaluate a per-token loss vector $\mathcal{L}(\x_{(j)})\in\mathbb{R}^{L}$ and compute the sequence-level NLL over non-padding tokens:
{\small
\begin{equation}
  \label{eq:supp:seq-nll}
  \widehat{\mathrm{NLL}}_{(j)} := \sum_{\ell = 1}^{L} a_{(j),\ell}\,\mathcal{L}(\x_{(j)})_{\ell}.
\end{equation}
\par
}
For diffusion models, $\widehat{\mathrm{NLL}}_{(j)}$ is computed via Monte Carlo, as described in \supp{supp:nll-pure-models} and \supp{supp:mc-elbo-blockgen}.
\paragraph{Dataset-level evaluation}
Let $N_{(j)} := \sum_{\ell=1}^{L} a_{(j),\ell}$ denote the number of non-padding tokens in sequence $j$, and let $N_{\mathrm{val}} := \sum_{j=1}^{J} N_{(j)}$. We compute the dataset-level perplexity as
{\small
\begin{equation}
  \label{eq:supp:dataset-nll}
  \widehat{\mathrm{NLL}}_{\mathrm{val}} := \sum_{j=1}^{J}\widehat{\mathrm{NLL}}_{(j)},\;
  \mathrm{PPL} := \exp\!\left(\tfrac{\widehat{\mathrm{NLL}}_{\mathrm{val}}}{N_{\mathrm{val}}}\right).
\end{equation}
\par
}
Thus, our reported perplexity is $\exp(\widehat{\mathrm{NLL}}_{\mathrm{val}}/N_{\mathrm{val}})$. The next paragraphs specify $\mathcal{L}$ for each model class.
\subsection{Computing the NLL for pure models}
\label{supp:nll-pure-models}
This subsection specifies $\mathcal{L}$ when using a single factorization (AR, MDM, or Duo).
\paragraph{Autoregressive models}
For AR models, we use the exact token-level negative log-likelihood
{\small
\begin{equation}
  \mathcal{L}_{\mathrm{AR}}(\x_{(j)})_{\ell} := -\log p_\theta(x_{(j),\ell}\mid x_{(j),<\ell}),
\end{equation}
\par
}
and set $\mathcal{L}(\x_{(j)}) := \mathcal{L}_{\mathrm{AR}}(\x_{(j)})$ in \Eqn{eq:supp:seq-nll}.
\paragraph{Masked diffusion (MDM)}
For masked diffusion \citep{sahoo2024simpleeffectivemaskeddiffusion, shi2025simplifiedgeneralizedmaskeddiffusion, ou2025absorbingdiscretediffusionsecretly}, the negative ELBO (NELBO) yields an NLL bound of the form
{\small
\begin{align}
  \label{eq:supp:mdm-nll}
  \mathrm{NELBO}_{\text{MDM}}(\x)
  :={}& -\mathbb{E}_{t,\mathbf{z}_t}\!\left[
    \tfrac{\alpha_t'}{1-\alpha_t}\sum_{\ell=1}^{L}
    \log \langle \hat{\x}_\theta^\ell(\mathbf{z}_t,t), \x^\ell \rangle
  \right] \nonumber \\
  \approx{}& -\tfrac{\alpha_{\tilde t}'}{1-\alpha_{\tilde t}}\sum_{\ell=1}^{L}
  \log \langle \hat{\x}_\theta^\ell(\mathbf{z}_{\tilde t},\tilde t), \x^\ell \rangle,
\end{align}
\par
}
which corresponds to a weighted cross-entropy loss over masked positions, with $t \sim \mathcal{U}[0,1]$ and $\mathbf{z}_t \sim q_t(\cdot \mid \x;\alpha_t)$ inside the expectation, and $\tilde t \sim \mathcal{U}[0,1]$, $\mathbf{z}_{\tilde t} \sim q_{\tilde t}(\cdot \mid \x;\alpha_{\tilde t})$ for the Monte Carlo estimate. Consistent with MDLM \citep{sahoo2024simpleeffectivemaskeddiffusion}, we use a single sample per sequence.
\paragraph{Uniform-state diffusion (Duo)}
For uniform-state diffusion \citep{schiff2025simpleguidancemechanismsdiscrete, sahoo2025diffusionduality}, the NELBO reads as
{\small
\begin{align}
  \label{eq:supp:duo-nll}
  &\mathrm{NELBO}_{\text{USDM}}(\x) \nonumber \\
  &:= \mathbb{E}_{t,\mathbf{z}_t}\!\left[\sum_{\ell=1}^{L} f_{\text{ELBO}}\!\left(\mathbf{z}_t^\ell,\hat{\x}_\theta^\ell(\mathbf{z}_t,t),\alpha_t;\x^\ell\right)\right] \nonumber \\
  &\approx \sum_{\ell=1}^{L} f_{\text{ELBO}}\!\left(\mathbf{z}_{\tilde t}^\ell,\hat{\x}_\theta^\ell(\mathbf{z}_{\tilde t},\tilde t),\alpha_{\tilde t};\x^\ell\right),
\end{align}
\par
}
with per-token term
{\small
\begin{align}
  \label{eq:supp:duo-felbo}
 f_{\text{ELBO}}&(\mathbf{z}_t,\hat{\x},\alpha_t;\x) := -\tfrac{\alpha_t'}{|\mathcal{V}|\alpha_t} \times \nonumber \\
  &\left[
    \tfrac{|\mathcal{V}|}{\barx_i} - \tfrac{|\mathcal{V}|}{(\barx_\theta)_i}
    - \sum_{r} \tfrac{\barx_r}{\barx_i}\log\tfrac{(\barx_\theta)_i\,\barx_r}{(\barx_\theta)_r\,\barx_i}
  \right],
\end{align}
\par
}
where $\barx = |\mathcal{V}|\alpha_t\x + (1-\alpha_t)\mathbf{1}$, $\barx_\theta = |\mathcal{V}|\alpha_t\hat{\x} + (1-\alpha_t)\mathbf{1}$, and $i=\arg\max_{r}(\mathbf{z}_t)_r$. Recall that $|\mathcal{V}|$ denotes the vocabulary size.
$\tilde t \sim \mathcal{U}[0,1]$ and $\mathbf{z}_{\tilde t} \sim q_{\tilde t}(\cdot \mid \x;\alpha_{\tilde t})$. Following MDLM \citep{sahoo2025diffusionduality}, we use a single Monte Carlo estimate per sequence.
\subsection{Computing the ELBO with BlockGen}
\label{supp:mc-elbo-blockgen}
\paragraph{ELBO computation for a single block size}
Fix a block size $s$ and a sequence $\x_{(j)}$. For each Monte Carlo draw, we sample $t \sim \mathcal{U}[0,1]$. We then sample $\mathbf{z}_t \sim q_t(\cdot \mid \x_{(j)};\alpha_t)$. Using the denoiser output and the chosen objective, we compute per-token losses $\mathcal{L}_{s}(\x_{(j)};t,\mathbf{z}_t)\in\mathbb{R}^{L}$. The corresponding sequence-level NLL is
{\small
\begin{equation}
  \mathrm{NLL}_{s,(j)}(t,\mathbf{z}_t) := \sum_{\ell=1}^{L} a_{(j),\ell}\,\mathcal{L}_{s}(\x_{(j)};t,\mathbf{z}_t)_{\ell}.
\end{equation}
\par
}
We estimate the expected sequence NLL with $K_{\mathrm{MC}}{=}8$ Monte Carlo draws:
{\small
\begin{equation}
  \label{eq:supp:mc-nll}
  \widehat{\mathrm{NLL}}_{s,(j)}
  := \frac{1}{K_{\mathrm{MC}}}\sum_{k=1}^{K_{\mathrm{MC}}}\mathrm{NLL}_{s,(j)}(t_k,\mathbf{z}_{t_k}).
\end{equation}
\par
}
We use $K_{\mathrm{MC}}{=}8$, and found it stable across seeds.
\paragraph{Evaluation of the BlockGen ELBO}
For the BlockGen parameterization \Eqn{eq:blockgen-mixture} with block sizes $\{s_i\}_{i=1}^{M}$ and weights $\boldsymbol{\gamma}$, we first compute $\widehat{\mathrm{NLL}}_{s_i,(j)}$ for each $i$ with $\gamma_i>0$ using \Eqn{eq:supp:mc-nll}. For evaluation, we always report the log-sum-exp (LSE) bound
{\small
\begin{equation}
  \label{eq:supp:valid-lse}
  \widehat{\mathrm{NLL}}_{\mathrm{LSE},(j)}
  := -\log\sum_{i:\gamma_i>0}\gamma_i \exp\!\left(-\widehat{\mathrm{NLL}}_{s_i,(j)}\right),
\end{equation}
\par
}
and set $\widehat{\mathrm{NLL}}_{(j)} := \widehat{\mathrm{NLL}}_{\mathrm{LSE},(j)}$ in \Eqn{eq:supp:dataset-nll}. We finally average over the dataset and exponentiate to obtain the final ELBO.
\section{Bounds on the Likelihood}
\label{app:proofs}
\subsection{BlockGen ELBOs}
\label{app:blockgen-elbo-proof}
Recall the definition of the BlockGen mixture density from \Eqn{eq:blockgen-mixture}:
{\small
\begin{equation}
  p_\theta^{\text{BlockGen}}(\x) = \sum_{i=1}^M \gamma_i \, p_\theta^{(s_i)}(\x),
\end{equation}
\par
}
where $\boldsymbol{\gamma} \in \Delta^M$ represents the mixture weights and each $p_\theta^{(s_i)}$ is a valid density. We assume each component $i$ admits a tractable lower bound $\mathcal{E}^{(s_i)}(\theta, \x)$ such that:
{\small
\begin{align}
  \log p_\theta^{(s_i)}(\x) &\geq \mathcal{E}^{(s_i)}(\theta, \x), \nonumber \\
  \text{so}\quad p_\theta^{(s_i)}(\x) &\geq e^{\mathcal{E}^{(s_i)}(\theta, \x)}.
\end{align}
\par
}
Substituting this inequality into the mixture definition:
{\small
\begin{equation}
  p_\theta^{\text{BlockGen}}(\x) = \sum_{i=1}^M \gamma_i \, p_\theta^{(s_i)}(\x) \geq \sum_{i=1}^M \gamma_i \, e^{\mathcal{E}^{(s_i)}(\theta, \x)}.
\end{equation}
\par
}
Taking the logarithm of both sides yields the \emph{log-sum-exp bound}:
{\small
\begin{align}
  \log p_\theta^{\text{BlockGen}}(\x) &\geq \log \left( \sum_{i=1}^M \gamma_i \, e^{\mathcal{E}^{(s_i)}(\theta, \x)} \right) \nonumber \\
  &= \log \left( \sum_{i=1}^M e^{\mathcal{E}^{(s_i)}(\theta, \x) + \log \gamma_i} \right).
\end{align}
\par
}
Finally, we can derive the \emph{mixture likelihood bound} using Jensen's inequality and the concavity of the $\log$:
{\small
\begin{align}
  \log \left( \sum_{i=1}^M \gamma_i \, e^{\mathcal{E}^{(s_i)}(\theta, \x)} \right) &\geq \sum_{i=1}^M \gamma_i \log \left( e^{\mathcal{E}^{(s_i)}(\theta, \x)} \right) \nonumber \\
  &= \sum_{i=1}^M \gamma_i \, \mathcal{E}^{(s_i)}(\theta, \x).
\end{align}
\par
}
\subsection{Alternative Parameterization via Geometric Mean}
\label{app:geometric-mean-proof}
We show that parameterizing the generative model with a geometric mean of component densities, denoted as $p_\theta^{\text{GMP}}$ also admits the mixture likelihood bound:
{\small
\begin{align}
  p_\theta^{\text{GMP}}(\x) &= \frac{1}{Z_\theta} \prod_{i=1}^M \left( p_\theta^{(s_i)}(\x) \right)^{\gamma_i}, \\
  \text{where}\quad Z_\theta &= \sum_{\x' \in \mathcal{V}^L} \prod_{i=1}^M \left( p_\theta^{(s_i)}(\x') \right)^{\gamma_i}. \nonumber
\end{align}
\par
}
The log-likelihood is given by:
{\small
\begin{equation}
  \label{eqn:gmp-log-likelihood}
  \log p_\theta^{\text{GMP}}(\x) = \sum_{i=1}^M \gamma_i \log p_\theta^{(s_i)}(\x) - \log Z_\theta.
\end{equation}
\par
}
To lower-bound \Eqn{eqn:gmp-log-likelihood}, we upper-bound the partition function $Z_\theta$.
For $p,q\in[1,\infty]$ with $1/p+1/q=1$, Hölder's inequality reads
{\small
\begin{equation}
  \sum_{\x'} |f(\x')g(\x')|
  \le \left(\sum_{\x'} |f(\x')|^p\right)^{\!1/p} \left(\sum_{\x'} |g(\x')|^q\right)^{\!1/q}.
\end{equation}
\par
}
Applying this inequality repeatedly yields the generalized form: for $p_i\in[1,\infty]$ with $\sum_i 1/p_i = 1$,
{\small
\begin{equation}
  \sum_{\x'} \prod_{i=1}^M |f_i(\x')|
  \le
  \prod_{i=1}^M \left(\sum_{\x'} |f_i(\x')|^{p_i}\right)^{1/p_i}.
\end{equation}
\par
}
Assume $\gamma_i>0$ (terms with $\gamma_i=0$ equal 1 and can be ignored), set $f_i(\x') := (p_\theta^{(s_i)}(\x'))^{\gamma_i}$ and choose $p_i := 1/\gamma_i$ so that $\sum_i 1/p_i = \sum_i \gamma_i = 1$. Applying the generalized form to $Z_\theta$ gives:
{\small
\begin{align}
  Z_\theta &= \sum_{\x'} \prod_{i=1}^M (p_\theta^{(s_i)}(\x'))^{\gamma_i} \nonumber \\
  &\leq \prod_{i=1}^M \left( \sum_{\x'} \left( (p_\theta^{(s_i)}(\x'))^{\gamma_i} \right)^{1/\gamma_i} \right)^{\!\gamma_i} \nonumber \\
  &= \prod_{i=1}^M \left( \sum_{\x'} p_\theta^{(s_i)}(\x') \right)^{\!\gamma_i}.
\end{align}
\par
}
Since each component $p_\theta^{(s_i)}$ is normalized, $\sum_{\x'} p_\theta^{(s_i)}(\x') = 1$. Therefore:
{\small
\begin{align}
  Z_\theta &\leq \prod_{i=1}^M (1)^{\gamma_i} = 1, \nonumber \\
  &\text{so}\;\; \log Z_\theta \leq 0 \;\;\text{and}\;\; -\log Z_\theta \geq 0.
\end{align}
\par
}
Substituting this back into the log-likelihood:
{\small
\begin{equation}
  \log p_\theta^{\text{GMP}}(\x) \geq \sum_{i=1}^M \gamma_i \log p_\theta^{(s_i)}(\x).
\end{equation}
\par
}
Finally, using the per-component tractable lower bound assumption $\log p_\theta^{(s_i)}(\x) \geq \mathcal{E}^{(s_i)}(\theta, \x)$:
{\small
\begin{equation}
  \log p_\theta^{\text{GMP}}(\x) \geq \sum_{i=1}^M \gamma_i \, \mathcal{E}^{(s_i)}(\theta, \x).
\end{equation}
\par
}
Thus, the Geometric Mean Parameterization also admits the \emph{Jensen bound} as a valid ELBO.
%


\newpage

\newpage

\newpage
\begin{table*}[t]
\centering
\caption{\textbf{Training cost.} All models use the $170$M-parameter backbone and train for $1$M steps with batch size $512$. On LM1B, we use 4 H100, on TinyGSM 8 H100, on OpenWebText, 16 H100.}
\label{tab:training-cost}
\footnotesize
\setlength{\tabcolsep}{6pt}
\renewcommand{\arraystretch}{1.1}
\begin{tabular}{l r r r r r r r r r}
\toprule
 & \multicolumn{3}{c}{Steps/sec} & \multicolumn{3}{c}{Duration (h)} & \multicolumn{3}{c}{GPU-hours} \\
\cmidrule(lr){2-4} \cmidrule(lr){5-7} \cmidrule(lr){8-10}
Model    & LM1B   & TinyGSM       & OWT    & LM1B   & TinyGSM       & OWT     & LM1B   & TinyGSM       & OWT      \\
\midrule
AR       & $12.56$ & $5.44$        & $5.96$  & $22.1$ & $51.1$        & $46.6$  & $88.5$  & $408.5$       & $745.7$  \\
MDLM     & $8.14$  & $3.04$        & $3.46$  & $34.1$ & $91.4$        & $80.3$  & $136.5$ & $731.0$       & $1284.5$ \\
Duo      & $8.14$  & $3.04$        & $3.46$  & $34.1$ & $91.4$        & $80.3$  & $136.5$ & $731.0$       & $1284.5$ \\
BlockGen & $5.53$  & $2.52$        & $2.45$  & $50.2$ & $110.2$       & $113.4$ & $200.9$ & $881.8$       & $1814.1$ \\
\bottomrule
\end{tabular}
\end{table*}

\newpage
\begin{table*}[t]
  \caption{Validation perplexity after 250k training steps for single-block-size models. Masked BDMs trained with cross-entropy (CE) matches or outperforms block diffusion trained with the ELBO \citep{shi2025demystifyingdiffusionobjectivesreweighted} and performs comparably to BD3-LM \citep{arriola2025blockdiffusioninterpolatingautoregressive} without their variance-reduction scheme. With uniform noise, CE improves perplexity only at small context sizes. LM1B (wrap) uses sentence packing; LM1B (no wrap) pads shorter sequences. All models are trained from scratch.}
  \label{tab:valid-ppl-after-250k-steps}
  \centering
  \footnotesize
  \setlength{\tabcolsep}{6pt}
  \renewcommand{\arraystretch}{1.1}
  {%
  \newcommand{\yes}{\(\circ\)}
  \newcommand{\no}{--}
  \newcommand{\tabrow}{\hspace*{0.8em}}
  \resizebox{\textwidth}{!}{%
  \begin{tabular}{@{}l cccc cccc cccccc@{}}
  \toprule
  & \multicolumn{4}{c}{LM1B (wrap)} & \multicolumn{4}{c}{LM1B (no wrap)} & \multicolumn{6}{c}{OWT} \\
  \cmidrule(lr){2-5} \cmidrule(lr){6-9} \cmidrule(lr){10-15}
  & \(L{=}1\) & \(L{=}4\) & \(L{=}16\) & \(L{=}128\) & \(L{=}1\) & \(L{=}4\) & \(L{=}16\) & \(L{=}128\)
  & \(L{=}1\) & \(L{=}4\) & \(L{=}16\) & \(L{=}32\) & \(L{=}128\) & \(L{=}1024\) \\
  \midrule

  \multicolumn{15}{@{}l}{\textit{Baselines}} \\
  \tabrow AR   & 22.4 & \no  & \no  & \no & 20.5 & \no  & \no  & \no & 17.0 & \no  & \no  & \no & \no & \no \\
  \tabrow MDLM {\footnotesize\citep{sahoo2024simpleeffectivemaskeddiffusion}} & \no  & \no  & \no & 33.7 & \no & \no & \no & 33.0 & \no & \no & \no & \no & \no & 24.9  \\
  \tabrow Duo {\footnotesize\citep{sahoo2025diffusionduality}}  & \no  & \no  & \no & 39.2 & \no & \no & \no & 38.3 & \no  & \no & \no & \no & \no & 27.5 \\
  \midrule

  \multicolumn{15}{@{}l}{\textit{Masked Block Diffusion}} \\
  \tabrow BD3-LM {\footnotesize\citep{arriola2025blockdiffusioninterpolatingautoregressive}} & 22.4 & 28.9 & 32.6 & \no & 20.5 & 30.3 & 32.3 & \no & 17.0 & 20.9 & 22.8 & \no & \no & \no \\
  \tabrow BDM (ELBO; ours)& \no & 30.1 & 33.6 & \no & \no & 31.2 & 34.5 & \no & \no & 21.8 & 23.7 & 24.1 & 24.5 & \no \\
  \tabrow BDM (CE; ours)  & 22.4 & 28.2 & 31.9 & \no & 20.5 & 27.3 & 31.0 & \no & 17.0 & 20.9 & 22.8 & 23.5 & 24.2 & \no \\
  \midrule

  \multicolumn{15}{@{}l}{\textit{Uniform Block Diffusion}} \\
  \tabrow BDM (ELBO; ours)& \no & 33.7 & 38.2 & \no & \no & 34.4 & 38.3 & \no & \no & 23.3 & 25.7 & 25.9 & 27.1 & \no \\
  \tabrow BDM (CE; ours)  & 22.7 & 31.9 & 36.9 & \no & 21.2 & 30.2 & 35.3 & \no & 17.0 & 22.9 & 26.6 & 28.4 & 28.7 & \no \\
  \bottomrule
  \end{tabular}%
  }
  }
\end{table*}

\newpage
\begin{table*}[t]
\centering
\caption{\textbf{Ancestral baseline, GSM8K accuracy (block size 16, single-block model).} NFE is per block. Each row is one NFE budget. Columns split by sampling temperature ($T{=}1$, $T{=}0.1$) and by noising prior (masked, uniform).}
\label{tab:ancestral_variants_b16}
\footnotesize
\setlength{\tabcolsep}{6pt}
\renewcommand{\arraystretch}{1.1}
\begin{tabular}{l r r r r}
\toprule
 & \multicolumn{2}{c}{T=1} & \multicolumn{2}{c}{T=0.1} \\
\cmidrule(lr){2-3} \cmidrule(lr){4-5}
NFE & Masked & Uniform & Masked & Uniform \\
\midrule
4 & 3.9 & 20.9 & 29.8 & 42.5 \\
8 & 14.1 & 31.8 & 42.9 & 52.2 \\
16 & 26.2 & 36.1 & 51.3 & 55.0 \\
32 & 32.8 & 38.4 & 56.3 & 56.1 \\
64 & 36.5 & 39.0 & 57.6 & 57.1 \\
\bottomrule
\end{tabular}
\end{table*}

\newpage
\begin{table*}[t]
\centering
\caption{\textbf{ARPC sweep, GSM8K accuracy (block size 16, mixture from \Eqn{eq:blockgen-mixture} with $\gamma_1 = 0.05$ and $\gamma_{16} = 0.95$).} NFE is per block. \emph{Guide-every} ($\text{GE}$) sets the cadence of predictor-corrector steps. $n_\text{warmup}$ is the number of initial ancestral-only steps. Total NFE per block is $N_\text{step}+\lfloor(N_\text{step}-1-n_\text{warmup})/\text{GE}\rfloor+1$. The best variant per (NFE, temperature, prior) is in \textbf{bold}, the second-best is \underline{underlined}.}
\label{tab:arpc_variants_b16}
\footnotesize
\setlength{\tabcolsep}{6pt}
\renewcommand{\arraystretch}{1.1}
\begin{tabular}{ccc r r r r}
\toprule
 &  &  & \multicolumn{2}{c}{T=1} & \multicolumn{2}{c}{T=0.1} \\
\cmidrule(lr){4-5} \cmidrule(lr){6-7}
$N_\text{step}$ & $\text{GE}$ & $n_\text{warmup}$ & Masked & Uniform & Masked & Uniform \\
\midrule
\multicolumn{7}{l}{\textit{NFE\,4}} \\
2 & 1 & 0 & 7.0 & 6.9 & 33.8 & 33.0 \\
3 & 2 & 1 & \underline{8.6} & \textbf{19.6} & \underline{36.4} & \underline{42.4} \\
3 & 3 & 0 & \textbf{13.3} & \underline{17.2} & \textbf{37.3} & \textbf{43.3} \\
\addlinespace[2pt]
\multicolumn{7}{l}{\textit{NFE\,8}} \\
4 & 1 & 0 & 32.8 & 33.7 & \underline{51.6} & 52.6 \\
5 & 2 & 0 & \underline{33.1} & \underline{36.8} & 51.4 & \underline{54.7} \\
7 & 2 & 5 & 19.4 & 34.9 & 45.2 & 51.9 \\
6 & 3 & 0 & \textbf{35.4} & \textbf{37.1} & \textbf{52.7} & \textbf{56.0} \\
\addlinespace[2pt]
\multicolumn{7}{l}{\textit{NFE\,16}} \\
8 & 1 & 0 & \textbf{54.0} & \textbf{50.9} & \underline{55.8} & 56.3 \\
11 & 2 & 1 & 46.1 & \underline{48.8} & 55.1 & 56.8 \\
14 & 2 & 10 & 32.0 & 39.9 & 55.1 & 56.4 \\
12 & 3 & 0 & \underline{47.9} & 47.4 & \textbf{57.5} & \textbf{57.8} \\
14 & 4 & 8 & 36.1 & 41.0 & 55.0 & \underline{56.9} \\
\addlinespace[2pt]
\multicolumn{7}{l}{\textit{NFE\,32}} \\
16 & 1 & 0 & \textbf{58.4} & \textbf{56.6} & 58.7 & 58.3 \\
21 & 2 & 0 & 53.6 & 52.3 & \textbf{59.1} & \underline{58.4} \\
28 & 2 & 20 & 40.9 & 43.6 & 57.1 & \textbf{59.0} \\
30 & 2 & 26 & 39.3 & 39.9 & 56.4 & 56.1 \\
24 & 3 & 0 & \underline{55.5} & \underline{53.0} & \underline{58.9} & 56.6 \\
30 & 4 & 24 & 40.0 & 42.5 & 56.8 & 57.2 \\
\addlinespace[2pt]
\multicolumn{7}{l}{\textit{NFE\,64}} \\
32 & 1 & 0 & \textbf{57.4} & \textbf{56.3} & 59.1 & \underline{58.5} \\
43 & 2 & 1 & \underline{54.9} & 55.2 & \textbf{59.4} & 58.2 \\
56 & 2 & 40 & 44.0 & 46.1 & \underline{59.2} & 57.8 \\
60 & 2 & 52 & 40.3 & 41.4 & 57.2 & 58.3 \\
62 & 2 & 58 & 39.3 & 40.0 & 58.5 & 56.8 \\
48 & 3 & 0 & 54.6 & \underline{55.7} & 58.6 & \textbf{59.1} \\
62 & 4 & 56 & 39.5 & 40.4 & 58.5 & 57.3 \\
\addlinespace[2pt]
\bottomrule
\end{tabular}
\end{table*}

\newpage
\begin{table*}[t]
\centering
\caption{\textbf{EIPC sweep, GSM8K accuracy (block size 16, single-block model).} NFE is per block. \emph{Guide-every} ($\text{GE}$) sets the cadence of predictor-corrector steps. $n_\text{warmup}$ is the number of initial ancestral-only steps. $N_\text{step}={}$NFE because guided steps reuse the denoiser output and cost nothing extra. The best variant per (NFE, temperature, prior) is in \textbf{bold}, the second-best is \underline{underlined}.}
\label{tab:eipc_variants_b16}
\footnotesize
\setlength{\tabcolsep}{6pt}
\renewcommand{\arraystretch}{1.1}
\begin{tabular}{cc r r r r}
\toprule
 &  & \multicolumn{2}{c}{T=1} & \multicolumn{2}{c}{T=0.1} \\
\cmidrule(lr){3-4} \cmidrule(lr){5-6}
$\text{GE}$ & $n_\text{warmup}$ & Masked & Uniform & Masked & Uniform \\
\midrule
\multicolumn{6}{l}{\textit{NFE\,4}} \\
2 & 0 & \textbf{5.6} & \textbf{22.2} & \textbf{34.9} & \textbf{41.9} \\
3 & 1 & \underline{4.6} & \underline{22.1} & \underline{30.3} & \underline{41.3} \\
\addlinespace[2pt]
\multicolumn{6}{l}{\textit{NFE\,8}} \\
2 & 6 & \underline{15.2} & \underline{31.4} & \underline{43.0} & \underline{54.3} \\
3 & 0 & \textbf{16.5} & \textbf{35.7} & \textbf{44.5} & \textbf{55.0} \\
\addlinespace[2pt]
\multicolumn{6}{l}{\textit{NFE\,16}} \\
2 & 12 & \underline{27.2} & \underline{37.4} & \underline{50.9} & \underline{55.3} \\
4 & 0 & \textbf{27.9} & \textbf{42.9} & \textbf{52.8} & \textbf{57.2} \\
\addlinespace[2pt]
\multicolumn{6}{l}{\textit{NFE\,32}} \\
2 & 20 & 32.9 & \underline{42.5} & 56.1 & 57.6 \\
2 & 26 & \underline{33.1} & 39.4 & \underline{56.2} & \textbf{57.8} \\
4 & 0 & \textbf{34.0} & \textbf{48.7} & \textbf{58.0} & \underline{57.8} \\
\addlinespace[2pt]
\multicolumn{6}{l}{\textit{NFE\,64}} \\
2 & 40 & 35.0 & \underline{42.2} & \underline{57.5} & \textbf{59.4} \\
2 & 52 & 35.1 & 40.0 & 57.5 & 58.4 \\
2 & 58 & \underline{35.3} & 38.2 & 57.5 & 57.6 \\
4 & 0 & \textbf{40.0} & \textbf{47.9} & \textbf{58.5} & \underline{58.6} \\
\addlinespace[2pt]
\bottomrule
\end{tabular}
\end{table*}

\newpage
\begin{table*}[t]
\centering
\caption{\textbf{Ancestral baseline, GSM8K accuracy (block size 32, single-block model).} NFE is per block. Each row is one NFE budget. Columns split by sampling temperature ($T{=}1$, $T{=}0.1$) and by noising prior (masked, uniform).}
\label{tab:ancestral_variants_b32}
\footnotesize
\setlength{\tabcolsep}{6pt}
\renewcommand{\arraystretch}{1.1}
\begin{tabular}{l r r r r}
\toprule
 & \multicolumn{2}{c}{T=1} & \multicolumn{2}{c}{T=0.1} \\
\cmidrule(lr){2-3} \cmidrule(lr){4-5}
NFE & Masked & Uniform & Masked & Uniform \\
\midrule
4 & 1.2 & 9.5 & 11.4 & 23.4 \\
8 & 5.6 & 18.2 & 22.7 & 39.6 \\
16 & 13.2 & 25.7 & 36.5 & 46.0 \\
32 & 20.3 & 29.3 & 44.2 & 49.9 \\
64 & 24.4 & 31.2 & 47.6 & 50.0 \\
128 & 28.1 & 31.4 & 50.3 & 49.6 \\
\bottomrule
\end{tabular}
\end{table*}

\newpage
\begin{table*}[t]
\centering
\caption{\textbf{ARPC sweep, GSM8K accuracy (block size 32, mixture from \Eqn{eq:blockgen-mixture} with $\gamma_1 = 0.05$ and $\gamma_{32} = 0.95$).} NFE is per block. \emph{Guide-every} ($\text{GE}$) sets the cadence of predictor-corrector steps. $n_\text{warmup}$ is the number of initial ancestral-only steps. Total NFE per block is $N_\text{step}+\lfloor(N_\text{step}-1-n_\text{warmup})/\text{GE}\rfloor+1$. The best variant per (NFE, temperature, prior) is in \textbf{bold}, the second-best is \underline{underlined}.}
\label{tab:arpc_variants_b32}
\footnotesize
\setlength{\tabcolsep}{6pt}
\renewcommand{\arraystretch}{1.1}
\begin{tabular}{ccc r r r r}
\toprule
 &  &  & \multicolumn{2}{c}{T=1} & \multicolumn{2}{c}{T=0.1} \\
\cmidrule(lr){4-5} \cmidrule(lr){6-7}
$N_\text{step}$ & $\text{GE}$ & $n_\text{warmup}$ & Masked & Uniform & Masked & Uniform \\
\midrule
\multicolumn{7}{l}{\textit{NFE\,4}} \\
2 & 1 & 0 & 2.0 & 1.5 & 15.5 & 15.2 \\
3 & 2 & 1 & \underline{2.5} & \underline{4.9} & \underline{18.2} & \underline{22.7} \\
3 & 3 & 0 & \textbf{5.5} & \textbf{7.4} & \textbf{19.3} & \textbf{24.3} \\
\addlinespace[2pt]
\multicolumn{7}{l}{\textit{NFE\,8}} \\
4 & 1 & 0 & \textbf{18.6} & 15.8 & \textbf{34.6} & 39.2 \\
5 & 2 & 0 & 16.0 & \textbf{18.3} & 32.8 & \underline{40.5} \\
7 & 2 & 5 & 8.1 & 15.2 & 29.0 & 37.3 \\
6 & 3 & 0 & \underline{17.6} & \underline{18.2} & \underline{33.5} & \textbf{41.0} \\
\addlinespace[2pt]
\multicolumn{7}{l}{\textit{NFE\,16}} \\
8 & 1 & 0 & \textbf{38.3} & \textbf{38.7} & \underline{42.7} & \textbf{47.8} \\
11 & 2 & 1 & 30.4 & 34.3 & 40.8 & 44.7 \\
14 & 2 & 10 & 19.3 & 26.7 & 42.5 & 43.7 \\
12 & 3 & 0 & \underline{32.5} & \underline{35.5} & \textbf{45.0} & \underline{46.3} \\
14 & 4 & 8 & 20.5 & 26.7 & 41.7 & 44.4 \\
\addlinespace[2pt]
\multicolumn{7}{l}{\textit{NFE\,32}} \\
16 & 1 & 0 & \textbf{49.6} & \textbf{48.0} & \textbf{48.7} & \underline{48.8} \\
21 & 2 & 0 & \underline{44.4} & \underline{45.4} & 46.0 & \textbf{49.9} \\
28 & 2 & 20 & 29.0 & 31.2 & 44.0 & 47.9 \\
30 & 2 & 26 & 23.7 & 28.3 & 44.8 & 47.4 \\
24 & 3 & 0 & 41.8 & 42.5 & \underline{46.5} & 48.0 \\
30 & 4 & 24 & 25.2 & 28.0 & 45.3 & 47.1 \\
\addlinespace[2pt]
\multicolumn{7}{l}{\textit{NFE\,64}} \\
32 & 1 & 0 & \textbf{51.3} & \textbf{52.8} & \underline{49.4} & 47.1 \\
43 & 2 & 1 & \underline{46.6} & \underline{50.0} & 48.1 & 46.7 \\
56 & 2 & 40 & 31.5 & 34.7 & 47.8 & \textbf{50.6} \\
60 & 2 & 52 & 29.4 & 31.3 & 48.6 & \underline{50.3} \\
62 & 2 & 58 & 27.7 & 29.3 & 48.1 & 49.6 \\
48 & 3 & 0 & 46.5 & 48.4 & \textbf{49.8} & 48.5 \\
62 & 4 & 56 & 27.0 & 29.9 & 48.3 & 49.6 \\
\addlinespace[2pt]
\multicolumn{7}{l}{\textit{NFE\,128}} \\
64 & 1 & 0 & \textbf{50.0} & \textbf{53.8} & 48.1 & 48.3 \\
85 & 2 & 0 & \underline{49.7} & \underline{52.1} & 49.4 & 48.2 \\
96 & 2 & 32 & 42.5 & 44.6 & 50.0 & 49.5 \\
112 & 2 & 80 & 33.6 & 34.5 & 49.2 & \underline{50.0} \\
120 & 2 & 104 & 30.6 & 32.3 & 50.0 & \textbf{50.3} \\
124 & 2 & 116 & 28.9 & 29.9 & \textbf{51.1} & 49.5 \\
126 & 2 & 122 & 28.2 & 30.9 & 49.4 & 49.8 \\
96 & 3 & 0 & 48.0 & 48.0 & \underline{50.4} & 47.6 \\
112 & 3 & 64 & 35.9 & 39.0 & 48.4 & 48.7 \\
126 & 4 & 120 & 27.8 & 30.2 & 49.4 & 49.5 \\
\addlinespace[2pt]
\bottomrule
\end{tabular}
\end{table*}

\newpage
\begin{table*}[t]
\centering
\caption{\textbf{EIPC sweep, GSM8K accuracy (block size 32, single-block model).} NFE is per block. \emph{Guide-every} ($\text{GE}$) sets the cadence of predictor-corrector steps. $n_\text{warmup}$ is the number of initial ancestral-only steps. $N_\text{step}={}$NFE because guided steps reuse the denoiser output and cost nothing extra. The best variant per (NFE, temperature, prior) is in \textbf{bold}, the second-best is \underline{underlined}.}
\label{tab:eipc_variants_b32}
\footnotesize
\setlength{\tabcolsep}{6pt}
\renewcommand{\arraystretch}{1.1}
\begin{tabular}{cc r r r r}
\toprule
 &  & \multicolumn{2}{c}{T=1} & \multicolumn{2}{c}{T=0.1} \\
\cmidrule(lr){3-4} \cmidrule(lr){5-6}
$\text{GE}$ & $n_\text{warmup}$ & Masked & Uniform & Masked & Uniform \\
\midrule
\multicolumn{6}{l}{\textit{NFE\,4}} \\
2 & 0 & \underline{0.8} & \underline{8.3} & \textbf{18.0} & \underline{25.7} \\
3 & 1 & \textbf{1.2} & \textbf{9.0} & \underline{10.8} & \textbf{26.9} \\
\addlinespace[2pt]
\multicolumn{6}{l}{\textit{NFE\,8}} \\
2 & 6 & \textbf{5.0} & \underline{15.9} & \underline{22.7} & \underline{39.6} \\
3 & 0 & \underline{4.8} & \textbf{22.8} & \textbf{29.9} & \textbf{43.9} \\
\addlinespace[2pt]
\multicolumn{6}{l}{\textit{NFE\,16}} \\
2 & 12 & \underline{13.0} & \underline{24.3} & \underline{36.2} & \underline{44.8} \\
4 & 0 & \textbf{14.6} & \textbf{34.0} & \textbf{40.4} & \textbf{52.1} \\
\addlinespace[2pt]
\multicolumn{6}{l}{\textit{NFE\,32}} \\
2 & 20 & \underline{20.1} & \underline{32.4} & 43.8 & \underline{50.8} \\
2 & 26 & 19.2 & 30.3 & \underline{44.2} & 48.1 \\
4 & 0 & \textbf{20.3} & \textbf{41.2} & \textbf{45.3} & \textbf{55.0} \\
\addlinespace[2pt]
\multicolumn{6}{l}{\textit{NFE\,64}} \\
2 & 40 & \underline{24.6} & \underline{39.0} & 47.2 & 51.4 \\
2 & 52 & \underline{24.6} & 32.7 & \underline{47.5} & 50.6 \\
2 & 58 & \underline{24.6} & 30.6 & \underline{47.5} & \underline{52.3} \\
4 & 0 & \textbf{27.4} & \textbf{45.8} & \textbf{49.4} & \textbf{55.3} \\
\addlinespace[2pt]
\multicolumn{6}{l}{\textit{NFE\,128}} \\
2 & 80 & \underline{28.6} & \underline{39.8} & 49.8 & \underline{53.4} \\
2 & 104 & 28.1 & 34.0 & 49.8 & 53.1 \\
2 & 116 & 28.0 & 30.6 & 50.0 & 51.7 \\
2 & 122 & 28.3 & 30.8 & \textbf{50.2} & 49.8 \\
4 & 0 & \textbf{29.8} & \textbf{48.1} & \underline{50.1} & \textbf{55.8} \\
\addlinespace[2pt]
\bottomrule
\end{tabular}
\end{table*}

\newpage
\begin{table*}[t]
\centering
\caption{\textbf{Single-block performance on TinyGSM at matched total NFE.} Both runs use 1024 total NFE per 512-token sequence (b16: 32 NFE/block $\times$ 32 blocks; b32: 64 NFE/block $\times$ 16 blocks). The best entry per column within each block is in \textbf{bold}. Entries marked ``--'' are pending.}
\label{tab:single_block_b16_vs_b32}
\footnotesize
\setlength{\tabcolsep}{6pt}
\renewcommand{\arraystretch}{1.1}
\begin{tabular}{l l l r r}
\toprule
 &  &  & \multicolumn{2}{c}{Accuracy (\%)} \\
\cmidrule(lr){4-5}
Block & Algo & Sampler & T=1 & T=0.1 \\
\midrule
16 & Masked & Ancestral & 32.8 & 56.3 \\
 & Masked & EIPC & 34.0 & \textbf{58.0} \\
 & Uniform & Ancestral & 38.4 & 56.1 \\
 & Uniform & EIPC & \textbf{48.7} & 57.8 \\
\midrule
32 & Masked & Ancestral & 24.4 & -- \\
 & Masked & EIPC & 27.4 & -- \\
 & Uniform & Ancestral & 31.2 & -- \\
 & Uniform & EIPC & \textbf{45.8} & -- \\
\bottomrule
\end{tabular}
\end{table*}

\end{document}